%% file: paper.tex
\documentclass[10pt]{article} 
\usepackage[preprint]{tmlr}

\input{math_commands.tex}

\usepackage{hyperref}
\usepackage{url}
\usepackage{amsmath,amssymb,amsfonts}
\usepackage{bm}
\DeclareMathSizes{8}{8}{7}{5}
\DeclareMathSizes{9}{9}{7}{5}
\usepackage{booktabs}
\usepackage{array}
\usepackage{graphicx}
\usepackage{float}
\usepackage{placeins}
\usepackage{xcolor}
 
\definecolor{revision}{RGB}{25,73,145}
\newif\ifrevisioncolor\revisioncolorfalse
\ifrevisioncolor\else\colorlet{revision}{black}\fi
\newcommand{\rev}[1]{{\color{revision}#1}}
\newenvironment{revision}{\begingroup\color{revision}}{\endgroup}
\hypersetup{hidelinks}

\newcommand{\Read}{\mathrm{Read}}

\newcommand{\LNf}{\mathrm{LN}_f}

\title{On the Tip of the Tongue:\\ Why LLMs Hallucinate Answers They Can Decode}

\renewcommand{\addr}{\normalfont\normalsize\itshape}
\author{\name Valeria Ruscio \\
\addr Intuition Machines Inc.
\AND 
\name Keiran Thompson \\
\addr Intuition Machines Inc.
}

\def\month{MM}  
\def\year{YYYY} 
\def\openreview{\url{https://openreview.net/forum?id=XXXX}} 

\begin{document}

\maketitle

\begin{abstract}
A language model can give the wrong answer even when the correct answer is decodable from its intermediate states. To study this gap between decodability and selection, we distinguish \emph{read} from \emph{write} at the first answer token. Read asks whether the gold token can be decoded from intermediate residual states under same-relation decoy controls. Write asks whether the final readout ranks that token first among content tokens. \rev{Under three different readers, with a randomized-label control,} a substantial fraction of failures remain readable \rev{while another content token is selected.} We explain this through the selection margin at the final readout, the difference between the answer logit and the logit of its strongest alternative, \rev{which is answer support minus alternative support, and can also be split into a context-averaged baseline linked to token frequency and an item-specific term.} \rev{Setting the answer support to the level typical of successful generations is sufficient to recover first-token selection for the majority of failures in most of the models we study; \rev{the original alternative remains ahead in most remaining failures under this edit, and this outcome follows directly from the readout geometry.}} \rev{Removing the frequency direction alone shifts selection but rarely recovers the answer.} \rev{Prompt variants of the same fact that succeed supply support that transfers to failing variants through the residual stream and through late MLP outputs, \rev{with less consistent effects through late attention.}} First-token recovery leaves most full answers wrong, \rev{which limits the recovery achieved by these edits} and separates three things that are easily conflated, decodability, recoverability, and generation.
\end{abstract}

\section{Introduction}
Large language models often fail on factual questions in a plausible way. Asked where a person was born, the model names the wrong city. Asked who wrote a book, it names another author. These errors are usually called factual hallucinations: fluent answers of the correct type that are false \citep{ji2023survey, huang2025survey}. This description leaves a mechanical question open. When the model gives the wrong answer, is the correct answer absent from the computation, or is it present but expressed too weakly at the final readout?

One way to approach this question is to compare generation with recognition. A model that fails to answer a question directly may still pick the correct answer from options, produce it for a paraphrased prompt, or assign it high probability under teacher forcing \citep{kadavath2022language,simhi2024distinguishing,gottesman2024estimating}. These cases are often described as a gap between what the model ``knows'' and what it says \citep{azaria2023internal,orgad2025llms}. We use this intuition as motivation, but it is unstable as a measurement target: reasonable recognition probes disagree about which facts count as available. If availability depends on the probe, then any mechanism proposed for the gap inherits the ambiguity of the probe.

We therefore replace the epistemic question with a mechanical one. On a single forward pass, we separate \emph{read} from \emph{write}. Read asks whether the first content token of the gold answer can be decoded from intermediate residual states under controlled probes. Write asks whether the final readout ranks that token first \rev{among content tokens}. This narrows the phenomenon: we study cases where evidence for the correct token is decodable inside the model, and the model still does not select that token.

\begin{revision}
Because the answer to ``is it readable'' depends on the reader, we do not rely on one. We decode with the logit lens, with a tuned translator that corrects the known bias of the logit lens at intermediate depth, and with a probe trained to separate gold tokens from same-relation decoys, and we compare the trained probe with a control trained on randomized labels, so that readability is always a statement about decodability under a named reader rather than about belief or knowledge.
\end{revision}

The explanation we propose for readable-but-unselected failures is a selection-margin account. Whether the gold token is written depends on the difference between its logit and the logit of its strongest competitor, \rev{and that margin is answer support minus alternative support. It can also be decomposed into a context-averaged baseline, which we show is linked to token frequency, and an item-specific term} \citep{demeter2020stolen,stolfo2024confidence}. The decomposition is not only descriptive, \rev{since projections along the answer and alternative directions can be set in the final state} and the effect on selection measured, both offline and during generation on the same items. \rev{\rev{Given the state before any edit, the geometry identifies failures for which the specified answer-setting target leaves the original alternative ahead. This is an exact consequence of the edit, not an independent test of the original error's cause.}}

We then ask where the support comes from. For the same fact, \rev{one prompt format can succeed while another fails}, which holds the subject, relation, answer and answer frequency fixed and gives a successful donor state for every failing prompt. Transferring activations from the successful phrasing into the failing one, \rev{with different-fact, random-depth and band-matched controls,} shows which parts of the computation carry fact-specific support that survives the rest of the forward pass. Finally, \rev{we follow the median-calibrated directed interventions through full generation}, because recovering the first token and producing the correct answer are different events, \rev{and the gap between them measures the remaining errors under these interventions.}

Our contribution is therefore not a new test of whether a model knows a fact, nor a complete account of multi-token answer generation, but a selection-margin account of one way factual errors arise at the first answer token, \rev{together with the controls that make the account testable: readability measured under three readers against a randomized control, interventions calibrated to the model's own successes and verified in generation, \rev{an algebraic criterion for failures that a specified answer-setting target cannot repair}, and transfers between prompt variants of the same fact. An earlier version of this work contained an extraction error at the final readout that distorted several of its numbers and one of its conclusions; Appendix~\ref{app:changes} lists what changed.}

\section{Related Work}
\label{sec:related}
Factual hallucination is often studied at the output level: the model produces fluent text that is unsupported, unverifiable, or false. TruthfulQA tests whether models repeat common falsehoods \citep{lin2022truthfulqa}. FActScore evaluates long generations by decomposing them into atomic claims \citep{min-etal-2023-factscore}. Black box methods such as SelfCheckGPT detect unreliable text through consistency across samples or queries \citep{manakul2023selfcheckgpt, agrawal2024do}. Theoretical work argues that calibrated models must hallucinate at some rate on rare facts \citep{kalai2024calibrated}. These methods describe when generated text is unreliable. We ask a more local question: in a failed factual generation, what happens to the correct answer token before the final readout selects another token?

A second line of work asks whether pretrained language models store factual associations in their parameters. LAMA queried masked language models with cloze prompts and treated them as implicit knowledge bases \citep{petroni-etal-2019-language}. PopQA and studies of long tail recall show that factual accuracy depends strongly on entity popularity and on the amount of pretraining evidence \citep{mallen-etal-2023-trust, kandpal2023large}. Work on eliciting latent knowledge and on hidden states related to hallucination asks whether activations contain information that the output does not express \citep{burns2023discovering, duan2024llmsknowhallucinationempirical, ji2024llm}. We share this motivation but use a narrower target at the token level: decodability of the gold first token at intermediate layers versus its rank at the final readout. Related interventions exploit internal signals at inference time: ITI shifts activations along truthful directions \citep{li2023inference}, DoLa contrasts intermediate and final layers to improve factuality \citep{chuang2024dola}, and late layers can override correct intermediate predictions \citep{halawi2024overthinking}.

Mechanistic studies of factual recall locate causal contributions in specific modules, layers, or editable representations. Knowledge neuron analyses link units to factual relations \citep{DBLP:journals/corr/abs-2104-08696}. Feed forward layers have been described as key value memories \citep{geva-etal-2021-transformer}, and they have been shown to build predictions by promoting concepts in vocabulary space \citep{geva2022transformer}; the recall of a factual association has been dissected into subject enrichment, relation propagation and attribute extraction \citep{geva2023dissecting}. Causal tracing and editing methods such as ROME and MEMIT find sites where factual associations can be modified \citep{meng2023locatingeditingfactualassociations, Meng2022MassEditingMI}. Our claim is complementary. We do not locate where a fact is stored. We analyze a later stage, in which a decodable answer token receives too little support at the readout, \rev{and we use activation transfers between prompt variants of the same fact to ask which parts of the computation contribute to the readout margin.}

Our read measure builds on the logit lens, which decodes intermediate residual states through the output map. Raw logit lens predictions can be biased at intermediate depth, and the tuned lens corrects this with layer specific translators \citep{belrose2025elicitinglatentpredictionstransformers}. We therefore treat read as decodability under specified probes, \rev{and we compare hard-decoy readability under the logit lens, under a fitted affine translator, and under a trained contrastive probe with a randomized-label control.} The margin analysis also connects to work on residual stream geometry, direct logit attribution, and frequency structure in the output space \citep{zhou2021frequencybaseddistortionscontextualizedword, oyama2023normwordembeddingencodes, cho2025understanding}. The unembedding turns residual directions into token logits, and directions linked to frequency can give common tokens baseline support.

\section{Final-Readout Geometry}
\label{sec:geometry}
We describe the read/write dissociation as a negative selection margin at the final readout. For item $i$, let $h_i=\LNf(h_i^{(L)})$ be the normalized final residual state at the answer position, and let $W_U[v]$ be the unembedding vector for vocabulary token $v$. The final logit of token $v$ is
\begin{equation}
    z_{i,v} = \langle h_i, W_U[v] \rangle .
\end{equation}
\rev{For the token-level experiments, let $\mathcal C$ be the set of non-special tokens \rev{whose decoded text contains an ASCII letter or digit}, the content tokens. For the gold answer token $a_i$, define the strongest content alternative}
\begin{equation}
    b_i = \arg\max_{v\in\mathcal C\setminus\{a_i\}} z_{i,v},
\end{equation}
and the selection margin
\begin{equation}
    M_i(a_i) = z_{i,a_i} - z_{i,b_i}.
\end{equation}
\rev{A positive margin means that the gold token is selected among content tokens, and a negative margin means that another content token receives more support at the readout. Unrestricted generation, in which the model is free to emit a non-content token first, is evaluated separately.}

This separates two quantities that are often conflated. The first is whether evidence for the answer can be decoded before the final readout. The second is whether the final residual state gives the answer enough support to rank first. Read measures the first quantity; write measures the second. An answer that is readable but not selected is not a paradox. \rev{It is an answer with measurable intermediate decodability and a negative final margin, or a tie resolved in favor of another token.}

For the gold token, the relevant quantity at the readout is the projection of the residual state onto the answer's unembedding direction. Let
\begin{equation}
    u_{a_i} = \frac{W_U[a_i]}{\|W_U[a_i]\|}
\end{equation}
be the unit unembedding direction of the answer token. Then
\begin{equation}
    z_{i,a_i} = \langle h_i,u_{a_i}\rangle \|W_U[a_i]\|.
\end{equation}
The answer logit is the product of a projection term, which depends on the state, and a norm term, which is fixed per token. We call
\begin{equation}
    \pi_i(a_i) = \langle h_i,u_{a_i}\rangle
\end{equation}
the answer direction support. \rev{For a fixed token, changes in its logit come from the projection; selection also depends on competing logits.}

\begin{revision}
This geometry defines a family of interventions and, with them, a prediction. Setting the answer direction support of a failed item to a target $T$, which we draw from the distribution of $\pi$ among the model's own successful generations, replaces $h_i$ by $h_i + (T-\pi_i(a_i))\,u_{a_i}$. The answer logit becomes exactly $T\|W_U[a_i]\|$, and the logit of the original alternative moves by $(T-\pi_i(a_i))\langle u_{a_i}, W_U[b_i]\rangle$, since the two unembedding directions are not orthogonal. The alternative therefore keeps an excess of support after the edit equal to
\begin{equation}
    e_i = z_{i,b_i} + (T-\pi_i(a_i))\langle u_{a_i}, W_U[b_i]\rangle - T\|W_U[a_i]\| ,
    \label{eq:excess}
\end{equation}
and in exact arithmetic the original alternative still beats the answer if and only if $e_i>0$. \rev{The sign of $e_i$ is computable before applying the edit. Positive excess certifies failure at this target; nonpositive excess does not guarantee recovery, since a third token or a tie may prevent selection.}
\end{revision}

Answer direction support does not determine the whole margin. To expose the readout baseline, split the final residual state into a mean over items and a deviation per item:
\begin{equation}
    h_i = \bar h + \Delta h_i, \qquad \bar h = \mathbb{E}_i[h_i].
\end{equation}
Each logit then decomposes into a baseline term and a contextual term:
\begin{equation}
    z_{i,v} = \underbrace{\langle \bar h,W_U[v]\rangle}_{b_v} + \underbrace{\langle \Delta h_i,W_U[v]\rangle}_{c_{i,v}},
\end{equation}
and the selection margin decomposes exactly as
\begin{equation}
    M_i(a_i) = \underbrace{(b_{a_i}-b_{b_i})}_{M_i^{base}} + \underbrace{(c_{i,a_i}-c_{i,b_i})}_{M_i^{ctx}}.
\end{equation}
The baseline term is the readout preference of the model under the mean final residual state for this prompt distribution, and the contextual term is the deviation associated with the specific prompt. Because $\bar h$ is estimated on the evaluation distribution, the baseline describes the model under this distribution; it is not a prior that is independent of the data, \rev{and we therefore also estimate it on neutral text and on the failures alone.} A negative baseline margin means that the reference state already favors the token that was selected; since the alternative is chosen from the item's own logits, the comparison remains conditioned on that competitor.

Frequency shapes this baseline. Suppose token frequency is partly carried by a shared direction $r$ in the unembedding:
\begin{equation}
    W_U[v] = \beta_v r + s_v,
\end{equation}
where $\beta_v$ correlates with token log frequency and $s_v$ is the remaining token specific part. Baseline logits then contain a frequency term whenever $\bar h$ has a component along $r$. A frequent alternative of the same type can therefore start with support from the baseline, and the answer signal must overcome both this offset and any item specific support for the alternative.

\rev{This account describes factual errors as failures of selection at the first token; the final-state edits do not by themselves identify their upstream cause.} The model can hold decodable, fact specific evidence for the correct token, but generation depends on how strongly that evidence appears at the final readout. When it appears too weakly relative to the competitor, the selected token can be a plausible alternative: close to the expected answer type, and false.

\section{Methodology}
\label{sec:methodology}
We evaluate frozen autoregressive transformer language models from the Llama, Qwen, and Mistral families, including base and instruction tuned variants, on PopQA subject--relation--object triples with answer aliases \citep{mallen-etal-2023-trust}\rev{; the main tables cover the eight checkpoints of the original study, between 1B and 8B parameters, and Appendix~\ref{app:more} repeats the entire protocol on five Qwen3 base checkpoints from 0.6B to 14B parameters. Each run uses 500 evaluation records and a separate pool of 1,500 records for fitting readers, drawn with a fixed seed after excluding records without aliases, with model weights in bfloat16. Prompts use a raw question-answering format rather than a chat template, with three fixed demonstrations drawn from records outside both splits prepended to every prompt; under zero-shot prompting the base checkpoints frequently begin their answer with a narrating token rather than the answer itself, which moves the answer out of the first generated position and makes the readout measurement ill defined. We therefore gate every run on a protocol check, the fraction of fully correct generations whose first generated token is an accepted alias token, and require it to exceed 70\%; \rev{the measured rates are 96.3--100\% for the eight main checkpoints and 94.7--100\% for Qwen3} (Appendix~\ref{app:methods}). A greedy response is marked correct if its normalized text contains a gold alias at word boundaries, after lowercasing and removing articles, punctuation and extra whitespace, and an empty prediction is never correct.}

\begin{revision}
All token-level analyses use a canonical target, the first token obtained by jointly tokenizing the prompt followed by a space and the first gold alias, so that the boundary variant is the one the tokenizer actually produces in that position. A record is a \emph{failure} when this target is not the highest-scoring content token at the final readout, and a \emph{success} when the full answer is correct and the first generated token is the canonical target; the latter population calibrates the interventions. These populations differ from full-answer failures and successes, since an accepted alias can begin with a different token, and Appendix~\ref{app:methods} reports all four counts per model.
\end{revision}

For readout analyses, we run a single forward pass on the prompt and inspect the residual stream at the final prompt position, immediately before the model emits its answer. Let $h_i^{(\ell)}$ denote the residual state for item $i$ at layer $\ell$. At each intermediate layer we apply the model's final normalization and unembedding, $z_i^{(\ell)}=W_U\LNf(h_i^{(\ell)})$. \rev{At the final layer we capture the residual immediately before the final normalization and normalize it once; the last entry of the hidden states returned by common implementations is already normalized, and normalizing it a second time, as an earlier version of this work did, distorts every final-readout quantity. Reconstructing the model's own logits from the corrected state gives 96--100\% top-1 agreement with the native forward pass on 50 records per model, \rev{against 16--88\% for the doubly normalized state; in the additional Qwen3 runs}, the maximum absolute logit error between the two paths is 0.06--0.12 and every remaining top-1 disagreement is an exact tie in the native logits (Appendices~\ref{app:methods} and~\ref{app:more}).}

Intermediate read is the strongest evidence for the gold token across the middle layers, $\Read^{\mathrm{int}}_i=\max_{\ell/L\in[0.4,0.9]}\log\mathrm{softmax}(z_i^{(\ell)})_{a_i}$, and hard-decoy readability requires this score to exceed the corresponding maximum for every decoy, where decoys are up to eight distinct first-token identities of other answers in the same PopQA relation, excluding the tokens of the item's own aliases. We also record the number of vocabulary tokens scoring strictly above the target at the final readout, which is the write rank. \rev{We repeat the read analysis with two fitted readers. The tuned reader fits, for each layer, an affine ridge translator from that layer's residual to the final pre-normalization residual, without any answer supervision. The contrastive probe learns a rank-128 bilinear score between the intermediate state and candidate input embeddings, trained to distinguish gold tokens from same-relation decoys, and a control probe with the same architecture, candidates and split is trained with a random candidate assigned as the positive. Both fitted readers use the 1,500-record pool, with 10\% held out for validation, and are evaluated only on the 500 evaluation records. Consensus readability is the intersection of the logit, tuned and trained-probe flags.}

Projection interventions operate on the normalized final residual. Answer setting (A) replaces the projection on $u_{a_i}$ with a target drawn from the model's own successes, the 25th, 50th or 75th percentile or the mean of $\pi$ among them, and the median is the main setting. Alternative setting (B) replaces the projection on the direction of the original alternative with the corresponding percentile of alternative support among successes; \rev{in the runs of Appendix~\ref{app:more} this edit is clamped so that it never raises an alternative that already sits below that level, written B$\downarrow$, \rev{which prevents an increase in that projection but does not guarantee higher recovery.}} The combined arm applies A and then B, and since the two directions are not orthogonal the second step \rev{can also move the answer projection.} Random controls displace the state by the norm of the A edit along a random unit direction. \rev{Every arm is evaluated offline, by decoding the edited state through the unembedding, and the three directed arms at the median target are also applied during generation, through a hook on the final normalization at the first decoding step, with non-content tokens masked at that step only and no further intervention; in the runs of Appendix~\ref{app:more} a mask-only control generates with the same first-step mask and no edit, which separates the behavioral effect of the mask from the effect of the edit. The same failures are used offline and in generation, and we report how often the offline and generated first tokens agree.}

For within-fact transfers we search, for each question, for one prompt variant whose first generated token is the canonical target with a correct full answer and another variant whose top content token is not the target, where variants change the question-answer formatting and instruction line but not the question. \rev{We then transfer last-position activations from the successful variant into the failing one and run the remaining computation: either the residual entering the late band, which begins at $\lfloor 0.75L\rfloor$, or the outputs of the late attention blocks only, or of the late MLP blocks only, or of every late block, \rev{and compare the donor arms against different-fact donors. The random-depth, early-band and middle-band controls match the number of components in the whole late band, not the attention-only or MLP-only arms.} Residual patching at the layer where the logit lens reads the gold token most strongly, and ablation of the gold read direction at that layer against a random direction of the same magnitude, complete the within-fact tests.}

For frequency analyses, token frequencies are estimated per tokenizer from a fixed WikiText-103 sample with add-one smoothing, the frequency direction is $r=\sum_{w\in\mathcal C}(f_w-\bar f)W_U[w]$ normalized to unit length, \rev{and the baseline is estimated from three references, all evaluation records, the failures alone, and neutral WikiText lines}. To test whether the frequency direction affects selection we remove the component of the final residual along $r$ and compare against random-direction ablations of the same norm.

\begin{revision}
Rates are reported with 95\% Wilson intervals, paired mean differences with bootstrap intervals over 2,000 resamples, binary paired comparisons with exact McNemar tests, and paired margin comparisons with Wilcoxon signed-rank tests, with Holm correction within each family of comparisons. All claims are reported per model, and Appendix~\ref{app:methods} specifies the statistical units.
\end{revision}

\section{Results}
\label{sec:results}

\subsection{Readable failures are usually not selected}
The hard-decoy criterion identifies readable failures in every model (Table~\ref{tab:readers}). Among failures, between 23.2\% and 41.8\% remain readable under the logit lens\rev{, between 33.3\% and 42.5\% under the tuned reader, and between 27.2\% and 30.8\% under the trained probe,} \rev{and these failures, by definition, do not select the gold token among content tokens at the final readout.} \rev{The randomized-label probe, trained with the same architecture, candidates and split, flags only 10.0--14.9\% of failures, and the trained probe exceeds it by 13.8--20.6 percentage points in every model, with all paired intervals above zero (Appendix~\ref{app:read}); \rev{the candidate sets and training procedure alone do not reproduce the trained probe's rate under this control.} The three readers agree on the population but only partly on the items, with Jaccard overlaps between the logit and tuned readable sets of 0.36--0.59 and consensus fractions of 9.8--16.0\%, which is why we treat readability as a property under a named reader and report every downstream stratification per reader.}

\begin{table}[t]\centering\color{revision}
\caption{Hard-decoy readable fractions among content-selection failures under three readers and the randomized-label control. Consensus requires logit, tuned and trained-probe readability. I denotes an instruction-tuned checkpoint; $n_F$ is the number of failures among 500 records. Intervals and overlaps are in Appendix~\ref{app:read}.}\label{tab:readers}
\small
\setlength{\tabcolsep}{4pt}
\begin{tabular}{lrrrrrr}
\toprule
Model & $n_F$ & Logit & Tuned & Probe & Shuffled & Consensus \\
\midrule
Llama-3.1-8B & 358 & 36.0\% & 42.5\% & 30.2\% & 12.6\% & 13.4\% \\
Llama-3.2-1B & 423 & 27.2\% & 38.5\% & 28.1\% & 13.9\% & 10.4\% \\
Llama-3.2-3B & 383 & 41.8\% & 41.5\% & 27.2\% & 13.3\% & 14.9\% \\
Llama-3.2-3B-I & 399 & 37.3\% & 39.3\% & 27.8\% & 11.8\% & 12.8\% \\
Qwen2.5-3B & 409 & 23.2\% & 33.3\% & 30.8\% & 11.2\% & 9.8\% \\
Qwen2.5-3B-I & 426 & 25.6\% & 37.8\% & 30.3\% & 10.6\% & 10.8\% \\
Qwen2.5-7B & 403 & 25.8\% & 40.2\% & 29.3\% & 14.9\% & 12.2\% \\
Mistral-7B & 369 & 38.5\% & 41.7\% & 30.6\% & 10.0\% & 16.0\% \\
\bottomrule\end{tabular}\end{table}

\subsection{Answer support at the level of successes is sufficient for most failures}
The intervention asks a counterfactual question about every failure, readable or not: if this item had the answer direction support that is typical of the model's own successful generations, would the gold token be selected? \rev{Table~\ref{tab:edits} answers it at the median target. Setting the answer support alone recovers first-token selection in 83.0--92.2\% of failures in every model, against 0.0--0.6\% for a random displacement of the same norm, \rev{and setting the alternative alone recovers only} 2.6--7.6\%. The combined edit adds 1.0--7.9 percentage points over answer setting alone, an increment that is significant after Holm correction in six of the eight models (Appendix~\ref{app:interventions}), \rev{and answer setting outperforms alternative setting throughout the sweep. The combined arm falls below answer setting at the upper-quartile target in two models (Appendix~\ref{app:interventions}).}}

\begin{revision}
The failures that answer setting does not repair are the ones the geometry says it cannot. The excess of equation~\ref{eq:excess}, computed from the unedited state, is positive for 7.3--16.4\% of failures at the median target, and it predicts the outcome of the edit almost perfectly: \rev{no item with positive excess recovers, and at most four items per model fail despite nonpositive excess, where a third token or a tie can prevent selection} (Table~\ref{tab:predictor}). \rev{The share of failures with positive excess depends on the model, cohort and calibration target.} In these eight checkpoints it is roughly one in ten, but on the five Qwen3 base checkpoints of Appendix~\ref{app:more} answer setting alone recovers 47.0\% of failures at 0.6B, 56.9\% at 1.7B, 75.1\% at 4B, 73.2\% at 8B and 85.4\% at 14B, the share of failures with positive excess falls correspondingly from 50.7\% to 14.3\%, the combined edit adds 4.4--20.3 points with every increment significant, and the excess criterion again has no false positives and at most ten misses per model. The earlier version of this work read a division of this kind as a split between model families on the basis of a distorted readout; measured correctly it is \rev{a per-item property of the specified edit, generally more common in the smaller Qwen3 checkpoints but not monotonic with scale.}
\end{revision}

We note what this result does not show. The edit is directed at the gold token, so it uses information the model did not express, and it is a statement about the sufficiency of answer support at the readout, \rev{not an answer-blind correction method or evidence of the dominant upstream cause. It also does not depend on readability}, which we turn to next.

\begin{table}[t]\centering\color{revision}
\caption{First-token recovery under median-calibrated projection edits, on all failures. A sets the answer support to the median of successes, B sets the alternative support to the median alternative support of successes, and A$\to$B applies both. Random matches the norm of A. Brackets are 95\% Wilson intervals.}\label{tab:edits}
\small
\setlength{\tabcolsep}{4pt}
\begin{tabular}{lrrrrr}
\toprule
Model & $n_F$ & A & B & A$\to$B & Random \\
\midrule
Llama-3.1-8B & 358 & 84.9\% [80.8\%, 88.3\%] & 7.3\% & 90.5\% [87.0\%, 93.1\%] & 0.6\% \\
Llama-3.2-1B & 423 & 83.0\% [79.1\%, 86.3\%] & 4.3\% & 85.6\% [81.9\%, 88.6\%] & 0.5\% \\
Llama-3.2-3B & 383 & 83.0\% [78.9\%, 86.5\%] & 5.2\% & 90.9\% [87.6\%, 93.4\%] & 0.3\% \\
Llama-3.2-3B-I & 399 & 92.2\% [89.2\%, 94.5\%] & 3.5\% & 96.7\% [94.5\%, 98.1\%] & 0.0\% \\
Qwen2.5-3B & 409 & 90.7\% [87.5\%, 93.2\%] & 2.9\% & 94.6\% [92.0\%, 96.4\%] & 0.5\% \\
Qwen2.5-3B-I & 426 & 90.6\% [87.5\%, 93.0\%] & 2.6\% & 95.3\% [92.9\%, 96.9\%] & 0.0\% \\
Qwen2.5-7B & 403 & 89.3\% [85.9\%, 92.0\%] & 3.2\% & 90.6\% [87.3\%, 93.1\%] & 0.2\% \\
Mistral-7B & 369 & 87.0\% [83.2\%, 90.0\%] & 7.6\% & 94.9\% [92.1\%, 96.7\%] & 0.0\% \\
\bottomrule\end{tabular}\end{table}

\begin{table}[t]\centering\color{revision}
\caption{The alternative-excess criterion of equation~\ref{eq:excess} at the median target. A positive excess predicts that answer setting alone fails; TP, FP, FN and TN count agreement between the prediction and the offline outcome. \rev{Switches count a third token winning after the edit, whether or not the original alternative also remains above the gold token; they need not equal FN.}}\label{tab:predictor}
\small
\setlength{\tabcolsep}{4pt}
\begin{tabular}{lrrrrrrr}
\toprule
Model & $n_F$ & TP & FP & FN & TN & Excess $>0$ & Switches \\
\midrule
Llama-3.1-8B & 358 & 54 & 0 & 0 & 304 & 15.1\% & 0.0\% \\
Llama-3.2-1B & 423 & 68 & 0 & 4 & 351 & 16.1\% & 3.3\% \\
Llama-3.2-3B & 383 & 63 & 0 & 2 & 318 & 16.4\% & 1.0\% \\
Llama-3.2-3B-I & 399 & 29 & 0 & 2 & 368 & 7.3\% & 0.5\% \\
Qwen2.5-3B & 409 & 37 & 0 & 1 & 371 & 9.0\% & 0.5\% \\
Qwen2.5-3B-I & 426 & 39 & 0 & 1 & 386 & 9.2\% & 0.7\% \\
Qwen2.5-7B & 403 & 39 & 0 & 4 & 360 & 9.7\% & 1.0\% \\
Mistral-7B & 369 & 48 & 0 & 0 & 321 & 13.0\% & 0.3\% \\
\bottomrule\end{tabular}\end{table}

\subsection{\rev{Readable failures are not easier to recover}}
\begin{revision}
If readability meant that the answer was almost written, readable failures should be the easy ones to recover. They are not. \rev{Stratifying the median answer-setting edit by reader gives recovery of 74--90\% on readable failures (67--91\% for consensus)} against 83--92\% on all failures, and when each readable failure is matched to an unreadable failure with the closest initial margin, the readable member of the pair recovers less often in all eight models under the logit reader, by 6.8--16.3 percentage points, with the interval excluding zero in five of them (Table~\ref{tab:matched}; \rev{reader-specific rates and logit/probe matches in Appendix~\ref{app:matched}}). \rev{The same matched pairs have smaller edits, by 0.8--5.7 units of norm with every interval below zero. Their answer projections therefore sit closer to the calibration target, but this distance alone does not imply a higher answer logit: token norms differ and margin matching is approximate. The comparison does not establish that stronger competitors explain the difference. These older subsets also depend on question-string joins (Appendix~\ref{app:methods}).} The Qwen3 runs, which record these quantities directly, \rev{show a related pattern: at approximately matched margin} the readable member of each pair has higher answer support in every model, \rev{by 0.74--1.69 units with every interval above zero}, its alternative-limited share is higher in every model (for example 50\% against 20\% at 4B and 67\% against 48\% at 1.7B), \rev{and its recovery point estimate is lower by 8--31 points, with intervals excluding zero in three of five models (Appendix~\ref{app:more}).}

This is the selection-margin account applied to its own measurement. Readability is a statement about the answer's evidence at intermediate depth, recoverability is a statement about the margin at the readout, \rev{and the two need not coincide: intermediate readability does not determine the final margin or the response to a calibrated edit.} The high overall recovery rate is accordingly not evidence for a mechanism specific to readable failures; it establishes the sufficiency of answer support on the whole cohort, and the existence of readable failures is established separately by the reader analysis.
\end{revision}

\begin{table}[t]\centering\color{revision}
\caption{Readable (R) versus unreadable (U) failures under the logit reader, matched one-to-one on the initial answer--alternative margin, at the median answer-setting edit. Differences are R minus U with bootstrap 95\% intervals; \rev{the norm difference is the difference in distance to the calibration target, not the minimum edit needed for recovery.}}\label{tab:matched}
\small
\setlength{\tabcolsep}{4pt}
\begin{tabular}{lrrrr}
\toprule
Model & $n$ & R / U recovery & Difference [95\% CI] & Edit-norm difference [95\% CI] \\
\midrule
Llama-3.1-8B & 129 & 75.2\% / 91.5\% & $-16.3$ [$-25.6$, $-7.8$] & $-2.05$ [$-2.62$, $-1.48$] \\
Llama-3.2-1B & 114 & 76.3\% / 83.3\% & $-7.0$ [$-18.4$, $3.5$] & $-0.79$ [$-1.31$, $-0.24$] \\
Llama-3.2-3B & 161 & 80.1\% / 87.0\% & $-6.8$ [$-14.9$, $1.2$] & $-1.55$ [$-1.90$, $-1.22$] \\
Llama-3.2-3B-I & 148 & 87.2\% / 96.6\% & $-9.5$ [$-15.5$, $-3.4$] & $-2.09$ [$-2.54$, $-1.65$] \\
Qwen2.5-3B & 95 & 83.2\% / 91.6\% & $-8.4$ [$-16.8$, $0.0$] & $-0.95$ [$-1.40$, $-0.47$] \\
Qwen2.5-3B-I & 108 & 85.2\% / 93.5\% & $-8.3$ [$-16.7$, $-0.9$] & $-1.71$ [$-2.59$, $-0.94$] \\
Qwen2.5-7B & 104 & 76.9\% / 92.3\% & $-15.4$ [$-24.0$, $-6.7$] & $-1.57$ [$-2.28$, $-0.91$] \\
Mistral-7B & 142 & 81.0\% / 91.5\% & $-10.6$ [$-18.3$, $-2.8$] & $-5.71$ [$-7.35$, $-4.16$] \\
\bottomrule\end{tabular}\end{table}

\subsection{A context-averaged baseline favors the selected alternative and is linked to frequency}
The baseline term of the margin decomposition favors the selected alternative in 66.9--82.1\% of failures, with a negative mean baseline margin in every model (Table~\ref{tab:baseline}); \rev{the mean final state favors the selected alternative. This is a context average, not a state before the prompt is processed.} \rev{This reference is estimated on all 500 evaluation records, successes included, and the result survives a change of reference: estimated on the failures alone or on neutral WikiText lines, the baseline is negative in 63.8--83.6\% of failures and its logits remain correlated with token log frequency, with correlations of 0.44--0.57 under every reference (Appendix~\ref{app:baseline}).}

\rev{The frequency-associated direction affects selection but its removal rarely recovers the answer.} Removing the component of the final state along the frequency direction shifts the selected token toward lower-frequency candidates and improves the gold margin relative to a random perturbation of the same norm, by 1.07 logits [0.62, 1.51] in Qwen2.5-3B and 0.29 [0.01, 0.58] in Mistral-7B for example, yet it recovers the gold token in only 1.7--5.4\% of failures, against 0.5--2.1\% for the random control and 83--92\% for answer setting. \rev{On the Qwen3 checkpoints the baseline is negative for 75--85\% of failures with the same frequency correlation, and removing the frequency direction recovers 0.7--1.1\% of failures, no more than the random control (Appendix~\ref{app:more}).} \rev{These edits act on a frequency-associated direction, without isolating frequency from other information it carries;} removing that advantage is not the same as supplying the support the answer lacks.

\begin{table}[t]\centering\color{revision}
\caption{All-record baseline evaluated on failures, and frequency-direction ablation. Negative baseline is the fraction of failures whose baseline margin favors the selected alternative; $r_{\mathrm{freq}}$ is the correlation between baseline logits and token log frequency; frequency recovery removes the frequency direction from the final state and random applies an equal-norm perturbation. Brackets are 95\% intervals.}\label{tab:baseline}
\small
\setlength{\tabcolsep}{4pt}
\begin{tabular}{lrrrrr}
\toprule
Model & Negative baseline & Mean margin & $r_{\mathrm{freq}}$ & Freq. recovery & Random \\
\midrule
Llama-3.1-8B & 67.0\% & $-1.242$ [$-1.526$, $-0.937$] & 0.567 & 2.5\% [1.3\%, 4.7\%] & 0.8\% \\
Llama-3.2-1B & 77.8\% & $-1.679$ [$-1.923$, $-1.421$] & 0.568 & 1.9\% [1.0\%, 3.7\%] & 1.4\% \\
Llama-3.2-3B & 71.3\% & $-1.335$ [$-1.577$, $-1.098$] & 0.570 & 3.4\% [2.0\%, 5.7\%] & 2.1\% \\
Llama-3.2-3B-I & 69.9\% & $-1.053$ [$-1.293$, $-0.810$] & 0.554 & 3.5\% [2.1\%, 5.8\%] & 0.5\% \\
Qwen2.5-3B & 76.8\% & $-1.828$ [$-2.099$, $-1.552$] & 0.561 & 2.2\% [1.2\%, 4.1\%] & 2.0\% \\
Qwen2.5-3B-I & 67.6\% & $-1.519$ [$-1.864$, $-1.170$] & 0.503 & 1.9\% [1.0\%, 3.7\%] & 0.9\% \\
Qwen2.5-7B & 82.1\% & $-2.026$ [$-2.274$, $-1.783$] & 0.495 & 1.7\% [0.8\%, 3.5\%] & 0.5\% \\
Mistral-7B & 66.9\% & $-1.141$ [$-1.408$, $-0.873$] & 0.500 & 5.4\% [3.5\%, 8.2\%] & 1.9\% \\
\bottomrule\end{tabular}\end{table}

\subsection{The same fact can be selected or not, and the difference transfers through the network}
Prompt variants hold everything about a fact fixed \rev{except the prompt formatting and instruction}, and across the eight models \rev{the search finds 18--45 record-level pairs} for which one variant produces the correct answer and another fails content selection. These pairs give a successful donor state for every failing prompt without changing the underlying association, and transferring activations between them asks which parts of the computation carry support that survives the rest of the forward pass.

\begin{revision}
Transferring the residual that enters the late band from the successful variant and running the remaining layers recovers first-token selection in 55.6--88.9\% of the failing variants, while the same transfer from a successful variant of a different question recovers 0--12.0\% (Table~\ref{tab:patching}), so the support that the late band receives is fact-specific. Within the late band, replacing only the MLP outputs with the donor's recovers 50.0--90.0\% against 0--9.7\% for a different-fact donor, with the paired difference significant after Holm correction in every model, whereas replacing only the attention outputs recovers 24.4--64.0\% against 0--35.5\% for a different-fact donor, a difference that is significant in two models only (Appendix~\ref{app:components}). Replacing the same number of components in the middle band of the network, leaving the late band free to respond, recovers 38.9--74.2\%, as much as the late MLPs in several models, and the early band recovers 20.6--45.0\%. \rev{These transfers show that middle-band and late-MLP outputs can carry useful support. They do not identify where that support first arises or whether MLPs amplify it; late attention also has a significant effect in two models. Different significance levels do not establish a difference between the MLP and attention effects.} The Qwen3 checkpoints, with 60 pairs each, give the same ordering: residual transfer recovers 66.7--90.0\% against 0--23.3\% for a different fact, late MLPs 56.7--75.0\% against 0--31.7\%, both significant in every model, and late attention 31.7--56.7\% against 15.0--23.3\%, significant in two of five (Appendix~\ref{app:more}). Replacing every late block at once, \rev{which fixes the patched last-position outputs throughout the late band}, recovers 64.7--100\%, but its margin over a random set of components sampled across all depths is not uniform, \rev{because that control changes component locations and the downstream computation left free to respond}; the band-matched controls are the informative ones.

\rev{Two further tests use the measured peak-read layer, or a middle-band fallback for a different failing prompt (Appendix~\ref{app:components}).} Patching the residual at that layer from the successful variant recovers the answer far more often than a different-fact donor, with differences of 28--85 points, and removing the gold read direction at that layer, in readable failures, lowers the final gold-versus-alternative margin by 1.4--3.5 logits where a random direction of the same magnitude changes it by less than 0.1 (Appendix~\ref{app:components}). \rev{The ablation changes the downstream margin, but the older readable-subset selection and peak assignments still require record-level verification (Appendix~\ref{app:methods}).}
\end{revision}

\begin{table}[t]\centering\color{revision}
\caption{Residual transfer at the entrance to the late band, from a successful prompt variant of the same question (same fact) or of a different question (different fact), followed by a forward pass through the remaining layers. \rev{$n$ counts record-level pairs for the arm rates. Paired tests use distinct questions: 44, 27, 31, 22, 18, 25, 20 and 33 in row order. Differences are percentage points with bootstrap 95\% intervals and Holm-adjusted exact McNemar $p$ values.}}\label{tab:patching}
\small
\setlength{\tabcolsep}{4pt}
\begin{tabular}{lrrrrr}
\toprule
Model & $n$ & Same fact & Different fact & Paired difference & $p_{\mathrm{Holm}}$ \\
\midrule
Llama-3.1-8B & 45 & 88.9\% & 2.2\% & 86.4\% [75.0\%, 95.5\%] & $<0.001$ \\
Llama-3.2-1B & 28 & 75.0\% & 0.0\% & 74.1\% [55.6\%, 88.9\%] & $<0.001$ \\
Llama-3.2-3B & 31 & 83.9\% & 0.0\% & 83.9\% [71.0\%, 96.8\%] & $<0.001$ \\
Llama-3.2-3B-I & 22 & 86.4\% & 0.0\% & 86.4\% [72.7\%, 100.0\%] & $<0.001$ \\
Qwen2.5-3B & 18 & 55.6\% & 5.6\% & 50.0\% [22.2\%, 77.8\%] & 0.059 \\
Qwen2.5-3B-I & 25 & 80.0\% & 12.0\% & 68.0\% [47.9\%, 88.0\%] & $<0.001$ \\
Qwen2.5-7B & 20 & 80.0\% & 0.0\% & 80.0\% [60.0\%, 95.0\%] & $<0.001$ \\
Mistral-7B & 34 & 67.6\% & 0.0\% & 69.7\% [54.5\%, 84.8\%] & $<0.001$ \\
\bottomrule\end{tabular}\end{table}

\subsection{Scope: first-token selection is not full-answer generation}
The interventions of Table~\ref{tab:edits} were also applied during generation, on the same failures, through a hook at the first decoding step. \rev{The generated first token agrees with the offline prediction in 97.4--100\% of items, \rev{so the offline and generated first-token outcomes closely agree, with occasional discrepancies.}} Answer setting recovers the generated canonical first token in 82.7--92.5\% of failures, \rev{and the full answer is correct in 30.5--38.0\% of the same failure cohort. The 13.8--28.0\% greedy accuracy on all records is a different denominator, not a matched baseline} (Table~\ref{tab:continuation}). The combined edit, here the symmetric A$\to$B arm, raises first-token recovery further but does not consistently raise full-answer correctness. \rev{Recovering the canonical first token is therefore far from sufficient and is not necessary when an accepted alias starts differently. Most recovered items still have incorrect full answers; some recovered tokens are shared with another entity, although that flag does not establish which entity the continuation names} (Appendix~\ref{app:continuation}). \rev{The first-step content mask on its own does not produce these answers: on the Qwen3 checkpoints, generating with the mask and no edit gives 1.5--5.0\% full-answer correctness on the same failures, against 20.7--37.1\% with the answer edit (Appendix~\ref{app:more}).} The mechanism studied in this paper governs how generation begins; the rest of the answer depends on later autoregressive states that a single readout edit does not control, \rev{which limits these particular edits without establishing an upper bound for every first-token intervention.}

\begin{table}[t]\centering\color{revision}
\caption{Greedy full-answer accuracy on all 500 records, and recovery under the median edits applied during generation on failures. First-token recovery uses the canonical token; full-answer correctness accepts any alias. Brackets are 95\% Wilson intervals.}\label{tab:continuation}
\small
\setlength{\tabcolsep}{4pt}
\begin{tabular}{lrrrrr}
\toprule
Model & Base full & A first & A full & A$\to$B first & A$\to$B full \\
\midrule
Llama-3.1-8B & 28.0\% & 84.6\% & 38.0\% [33.1\%, 43.1\%] & 90.5\% & 36.9\% [32.0\%, 42.0\%] \\
Llama-3.2-1B & 14.2\% & 82.7\% & 30.5\% [26.3\%, 35.0\%] & 85.3\% & 30.5\% [26.3\%, 35.0\%] \\
Llama-3.2-3B & 22.6\% & 82.8\% & 33.7\% [29.1\%, 38.6\%] & 90.9\% & 36.8\% [32.1\%, 41.8\%] \\
Llama-3.2-3B-I & 19.2\% & 92.5\% & 35.1\% [30.6\%, 39.9\%] & 97.0\% & 36.3\% [31.8\%, 41.2\%] \\
Qwen2.5-3B & 17.8\% & 90.5\% & 34.2\% [29.8\%, 39.0\%] & 94.4\% & 35.5\% [31.0\%, 40.2\%] \\
Qwen2.5-3B-I & 13.8\% & 89.9\% & 31.2\% [27.0\%, 35.8\%] & 94.8\% & 34.0\% [29.7\%, 38.7\%] \\
Qwen2.5-7B & 19.2\% & 89.8\% & 33.0\% [28.6\%, 37.7\%] & 91.1\% & 33.7\% [29.3\%, 38.5\%] \\
Mistral-7B & 26.8\% & 87.0\% & 33.9\% [29.2\%, 38.8\%] & 94.9\% & 36.3\% [31.6\%, 41.3\%] \\
\bottomrule\end{tabular}\end{table}

\subsection{Summary of the results}
The results support a layered account rather than a single explanation. \rev{The reader analysis establishes that intermediate decodability and final selection are separable, under three readers and against a randomized control, so the dissociation is more than type activation or an artifact of one probe. \rev{The calibrated intervention shows that setting answer support is sufficient for most failures in most models. The alternative-excess criterion identifies failures where this target leaves the original alternative ahead, a share that generally declines with Qwen3 scale but is not monotonic;} this replaces the division into model-level regimes that an earlier version of this work proposed on the basis of a distorted readout with a per-item criterion. \rev{The matched comparison shows no recovery advantage for readable failures; it does not establish stronger competitors as the cause.}} \rev{The baseline decomposition describes a readout preference:} the context-averaged readout leans toward the selected alternative and this lean is linked to frequency, \rev{yet removing the frequency-associated direction rarely recovers the answer.} \rev{The within-fact transfers show that the support is prompt-dependent and that it travels through the residual stream and the MLP outputs \rev{at the tested middle and late bands. Late attention also contributes in some models; the older read-direction ablations remain subject to the assignment checks in Appendix~\ref{app:methods}.}} The continuation analysis bounds the whole account to the first token, since recovering that token leaves most answers wrong.

\section{Limitations and Interpretation}
\label{sec:limitations}
The study covers one factual-recall distribution, PopQA, and one seed, with 500 evaluation records per model, on eight checkpoints between 1B and 8B parameters from three families \rev{and five Qwen3 base checkpoints from 0.6B to 14B}; a second factual distribution and checkpoints beyond 14B are left for future work. The within-fact transfers rest on 18--60 pairs per model and their intervals are correspondingly wide. \rev{Prompts are three-shot in a raw question-answering format, \rev{which the protocol gate shows usually places an accepted answer token first among fully correct generations}, but which need not be the most natural prompt for instruction-tuned models. Reader fitting and evaluation splits are disjoint at the record level, not at the level of subjects or questions, and in checkpoints with tied input and output embeddings the contrastive probe shares its candidate representations with the unembedding, so that it is a second fitted reader rather than a reader on an independent representation.}

The read measure should not be interpreted as a belief state. Later layers may transform, suppress or redirect intermediate information, and the three readers agree on the rate of readable failures while disagreeing on a substantial fraction of the items, so the central empirical claim is stated as decodability under named readers. \rev{The interventions use the gold direction and the within-fact transfers use a successful variant of the same question, so neither is an answer-blind correction method; \rev{they test what is sufficient at the readout and which patched sites can transfer support}, not how to obtain it without the answer.} The baseline is averaged over contexts and depends on the reference distribution, which is why three references are reported, and the transfers locate contributions to the readout margin, not the storage site of the association, since earlier layers may retrieve or compose the relevant information while later components determine how strongly it appears in the answer direction relative to the competitor. \rev{Finally, the first-step content mask used during patched generation differs from unrestricted generation, \rev{and the mask-only control measures that difference in the Qwen3 runs.}}

\section{Conclusion}
We studied factual hallucination through the narrower problem of selecting the first answer token, and we replaced the question of what a model knows by two mechanical ones, whether the gold token can be decoded from intermediate states under decoy controls, and whether the final readout gives it enough support to rank first. The two come apart in every model we studied, \rev{under three readers and against a randomized control}. \rev{A selection-margin analysis describes the gap: the margin is answer support minus competitor support, and can also be split into a frequency-linked baseline and an item-specific term,} \rev{answer support at the level of the model's own successes is sufficient for most failures in most models, \rev{positive excess identifies failures at the specified target and is generally more common in smaller Qwen3 checkpoints, and readable failures show no recovery advantage}}. Successful phrasings of the same fact supply support that transfers to failing ones \rev{through the residual stream and late MLP outputs}, which connects the readout account to the computation that precedes it. The account is deliberately scoped to the first token, and the continuation analysis shows why that scope matters, since recovering the first token leaves most answers wrong. What it isolates is one concrete way in which internally decodable factual evidence can fail to determine what the model says.

\newpage
\input{references.tex}

\newpage
\appendix

\begin{revision}
\section{Changes relative to the submitted version}
\label{app:changes}
\rev{The revised pipeline changes five aspects of the submitted implementation. The eight-model reports and the additional Qwen3 results below use the revised protocol, with further checks specific to Qwen3.} We list them here together with their consequences, since several numbers and one conclusion changed.

The final hidden state was normalized twice. The last entry of the hidden states returned by the model implementation is already the output of the final normalization, and the submitted code applied the normalization again before the unembedding. Reconstructing the model's own logits from that doubly normalized state gives top-1 agreement of only 16--88\% depending on the model, against 96--100\% for the corrected extraction (Table~\ref{tab:reconstruction}), so every final-readout quantity in the submitted version, the write rank, the margins, the answer and alternative support, the baseline decomposition and the interventions, was computed on a distorted state. The distortion was largest for Qwen2.5-3B, whose agreement was 16\%, \rev{and it undermines the submitted claim that answer setting almost never changes selection in that model;} \rev{under the revised protocol it recovers 90.7\% of failures, in line with the other checkpoints. Prompts and cohorts also changed, so this difference cannot be attributed solely to extraction,} and the division into answer-limited and alternative-limited models is withdrawn and replaced by the per-item criterion of equation~\ref{eq:excess}.

The interventions were applied to a pre-normalization state and evaluated offline only, on a population that mixed failures with other items, and without restricting the first token to content tokens. They are now applied to the normalized state at the exact point of the readout, on the failure cohort, both offline and during generation through a hook at the first decoding step, \rev{with a first-step content mask and, for Qwen3, a mask-only control. The offline and generated first tokens agree in 97.4--100\% of items in the eight-model runs and 97.0--99.3\% in Qwen3.}

The component patching was computed offline as a sum of component contributions rather than by running the network, and its random control was not matched in depth. Transfers are now forward passes with the donor activations in place, with different-fact, random-depth and band-matched controls, and with transfers of the residual stream, of late attention only and of late MLPs only in addition to the whole late band. As a result the submitted localization to late layers is replaced \rev{by evidence that middle-band and late-MLP transfers can recover selection, with late-attention effects in some models.}

The baseline $\bar h$ was estimated on the failures alone, which conditions the reference on the outcome. It is now estimated on all evaluation records, and the failure-only and neutral-text references are reported as sensitivity checks; the sign and the frequency association survive all three.

The answer matcher accepted a match in either direction and treated an empty prediction as a match. It is now one-directional and requires a non-empty normalized alias of at least two characters at word boundaries; the audit of label flips between the two matchers is in Table~\ref{tab:cohorts}.

Three further changes follow from the review. Readability is now measured under three readers with a randomized-label control instead of under the logit lens with a tuned-lens check on two models; the prompt protocol is three-shot with a protocol gate and a joint-tokenization target, because zero-shot prompting moves the answer out of the first generated position for several base checkpoints; and the analysis of frequency geometry over training checkpoints and in a toy model has been removed, since it is peripheral to the corrected claims. The whole protocol was also run on five additional checkpoints of a newer family, from 0.6B to 14B parameters (Appendix~\ref{app:more}). \rev{Rates and paired comparisons are reported with controls and uncertainty estimates; the remaining audit dependencies are described below.}
\end{revision}

\begin{revision}
\section{Additional Methodological Details}
\label{app:methods}
\paragraph{Models and data.}
The checkpoints are, under their Hugging Face identifiers, \texttt{Llama-3.1-8B}, \texttt{Llama-3.2-1B}, \texttt{Llama-3.2-3B} and \texttt{Llama-3.2-3B-Instruct} from \texttt{meta-llama}, \texttt{Qwen2.5-3B}, \texttt{Qwen2.5-3B-Instruct} and \texttt{Qwen2.5-7B} from \texttt{Qwen}, and \texttt{Mistral-7B-v0.1} from \texttt{mistralai}. PopQA records with empty aliases are excluded before a seed-0 shuffle; the first 500 records form the evaluation sample and the next 1,500 the reader-fitting pool, and the three demonstrations are drawn from records after both, choosing different relations and answers of at most 30 characters and three words. Relation pools for decoys use the canonical aliases of the complete shuffled dataset. No evaluation record was excluded for an invalid target token in any run. \rev{Table~\ref{tab:cohorts} gives the evaluation and calibration populations. The reports do not retain the total before alias filtering, exact demonstration identities \rev{or realized number of records visited during pair search}; these counts cannot be reconstructed from the aggregate tables.}

\paragraph{Prompts.}
The canonical template consists of the instruction line \texttt{Answer with a short factual answer.}, followed on the next line by \texttt{Question: \{q\}} and on the line after by \texttt{Answer:}, preceded by the three demonstrations in the same format. The within-fact search uses this template and three variants, \texttt{Q: \{q\}} followed by \texttt{A:} on the next line, \texttt{\{q\}} followed by \texttt{Answer:} on the next line, and \texttt{\{q\} The answer is} on one line, with the demonstrations reformatted to each; only the canonical format carries the instruction line. \rev{In the eight-model runs the search visits the first $\max(n_{\rm eval},4n_{\rm pairs})=500$ shuffled records, stops at the first successful and failing variants with the same canonical target token for each record, and is capped at 60 pairs. The Qwen3 search visits more records.}

\paragraph{Generation and matching.}
Generation is greedy with at most 16 new tokens and prompts are truncated at 512 tokens. Canonical targets are obtained by tokenizing the prompt followed by a space and the first alias and taking the first token beyond the prompt's own tokens; content tokens must contain an ASCII letter or digit and must not be special tokens. Unmodified generation is unrestricted; patched generation applies the content mask at its first step and then continues without masking or editing. Matching lowercases the text, removes articles, maps punctuation and non-alphanumeric characters to spaces and collapses whitespace, and a non-empty prediction is correct if it contains a normalized alias of at least two characters at word boundaries.

\begin{table}[!htbp]\centering\color{revision}
\caption{Evaluation and calibration populations. Full, canonical-first and any-alias-first accuracy use 500 records; $n_S$ counts successes (full answer correct and canonical first token), $n_P$ counts records whose first token is canonical but whose full answer is wrong, and flips counts records whose label differs between the submitted and the corrected matcher.}\label{tab:cohorts}
\footnotesize
\setlength{\tabcolsep}{4pt}
\begin{tabular}{lrrrrrr}
\toprule
Model & Full accuracy & Canonical & Any alias & $n_S$ & $n_P$ & Flips \\
\midrule
Llama-3.1-8B & 28.0\% [24.2\%, 32.1\%] & 28.4\% & 33.2\% & 124 & 18 & 10 \\
Llama-3.2-1B & 14.2\% [11.4\%, 17.5\%] & 15.4\% & 18.2\% & 62 & 15 & 10 \\
Llama-3.2-3B & 22.6\% [19.2\%, 26.5\%] & 23.4\% & 27.8\% & 101 & 16 & 12 \\
Llama-3.2-3B-I & 19.2\% [16.0\%, 22.9\%] & 20.2\% & 23.8\% & 81 & 20 & 16 \\
Qwen2.5-3B & 17.8\% [14.7\%, 21.4\%] & 18.2\% & 22.8\% & 76 & 15 & 13 \\
Qwen2.5-3B-I & 13.8\% [11.1\%, 17.1\%] & 15.0\% & 18.8\% & 57 & 18 & 16 \\
Qwen2.5-7B & 19.2\% [16.0\%, 22.9\%] & 19.4\% & 23.6\% & 82 & 15 & 13 \\
Mistral-7B & 26.8\% [23.1\%, 30.8\%] & 26.0\% & 30.4\% & 115 & 15 & 14 \\
\bottomrule\end{tabular}\end{table}

\paragraph{Extraction and reconstruction.}
The final residual is captured by a hook on the input of the final normalization and normalized once; the unembedding is applied in float32 to the normalized state. Table~\ref{tab:reconstruction} compares the resulting logits with the native forward pass on 50 records per model, for the corrected extraction and for the submitted one. The native logits come from the model's bfloat16 output path, \rev{so precision is a possible source of disagreement. The eight-model reports do not include per-record gaps and errors, and do not establish the cause of the remaining disagreements. Exact-tie diagnostics are reported only for Qwen3 in Appendix~\ref{app:more}.} The protocol gate, which is a separate check, is the fraction of fully correct generations whose first token is an accepted alias token.

\begin{table}[!htbp]\centering\color{revision}
\caption{Logit reconstruction on 50 records: normalization once (corrected) versus twice (submitted), top-1 agreement with the native forward pass and mean logit correlation, and the protocol gate.}\label{tab:reconstruction}
\footnotesize
\setlength{\tabcolsep}{4pt}
\begin{tabular}{lrrrrr}
\toprule
Model & Corrected top-1 & Submitted top-1 & Corrected corr. & Submitted corr. & Protocol gate [95\% CI] \\
\midrule
Llama-3.1-8B & 1.000 & 0.720 & 1.000 & 0.970 & 99.3\% [96.1\%, 99.9\%] \\
Llama-3.2-1B & 0.960 & 0.780 & 1.000 & 0.986 & 98.6\% [92.4\%, 99.8\%] \\
Llama-3.2-3B & 0.980 & 0.840 & 1.000 & 0.988 & 100.0\% [96.7\%, 100.0\%] \\
Llama-3.2-3B-I & 0.960 & 0.880 & 1.000 & 0.988 & 97.9\% [92.7\%, 99.4\%] \\
Qwen2.5-3B & 0.980 & 0.160 & 1.000 & 0.956 & 100.0\% [95.9\%, 100.0\%] \\
Qwen2.5-3B-I & 0.980 & 0.560 & 1.000 & 0.954 & 98.6\% [92.2\%, 99.7\%] \\
Qwen2.5-7B & 0.980 & 0.640 & 1.000 & 0.953 & 96.9\% [91.2\%, 98.9\%] \\
Mistral-7B & 0.980 & 0.840 & 1.000 & 0.971 & 96.3\% [91.6\%, 98.4\%] \\
\bottomrule\end{tabular}\end{table}

\paragraph{Reader fitting.}
The read band contains the layers $\ell$ with $0.4\leq\ell/L\leq0.9$. The tuned reader is an affine ridge regression from the residual at each band layer to the final pre-normalization residual, fitted in float64 with penalty $0.01\,\mathrm{tr}(X^\top X)/d$ on the weights and none on the intercept, and its held-out $R^2$ is reported in the run summaries. The contrastive probe scores a candidate $v$ at layer $\ell$ by $s_\ell(x,v)=(\mathrm{RMS}(x)U_\ell)^\top(E[v]V_\ell)$ with $U_\ell,V_\ell\in\mathbb R^{d\times128}$ and $E$ the input embedding matrix, trained with AdamW at learning rate $10^{-3}$, weight decay $10^{-4}$, batches of 64 and 30 epochs, keeping the epoch with the lowest validation loss; the randomized-label probe shares the split and the candidate sets and takes a uniformly random candidate as the positive. Both readers hold out 10\% of the eligible fitting records. Hard readability compares the maximum over the band of the gold score with the maximum over the band of each decoy's score, and the two maxima may occur at different layers. Probe scores are not calibrated across layers, and in checkpoints with tied embeddings the probe's candidate representations coincide with the unembedding rows.

\paragraph{Statistical units.}
Every rate is computed per evaluation record. In the eight-model runs the paired comparisons, the reader-stratified and matched subsets and the peak-layer lookups were keyed on the question string; \rev{repeated questions can collapse rows or assign flags and layers from another record. The reported small number of repeats does not bound their effect on matched subsets or the smaller patching and ablation cohorts. Paired tests retain the first row per question; Tables~\ref{tab:patchcomparisons} and~\ref{tab:peak} report these question counts. The affected older subgroup, paired and ablation analyses remain provisional until record-level assignments can be checked.} The Qwen3 runs of Appendix~\ref{app:more} key every join on a stable record identifier. Bootstrap intervals are over records or over pairs and do not include variation across seeds or across random-donor draws.
\end{revision}

\section{Reader Validation}
\label{app:read}
\begin{table}[!htbp]\centering\color{revision}
\caption{Reader-specific hard-decoy readable fractions among failures, with 95\% Wilson intervals.}\label{tab:readerci}
\footnotesize\setlength{\tabcolsep}{3pt}
\begin{tabular}{lrrrrr}\toprule
Model & $n_F$ & Logit & Tuned & Probe & Consensus \\
\midrule
Llama-3.1-8B & 358 & 36.0\% [31.2\%, 41.1\%] & 42.5\% [37.4\%, 47.6\%] & 30.2\% [25.6\%, 35.1\%] & 13.4\% [10.3\%, 17.3\%] \\
Llama-3.2-1B & 423 & 27.2\% [23.2\%, 31.6\%] & 38.5\% [34.0\%, 43.3\%] & 28.1\% [24.1\%, 32.6\%] & 10.4\% [7.8\%, 13.7\%] \\
Llama-3.2-3B & 383 & 41.8\% [36.9\%, 46.8\%] & 41.5\% [36.7\%, 46.5\%] & 27.2\% [22.9\%, 31.8\%] & 14.9\% [11.7\%, 18.8\%] \\
Llama-3.2-3B-I & 399 & 37.3\% [32.7\%, 42.2\%] & 39.3\% [34.7\%, 44.2\%] & 27.8\% [23.6\%, 32.4\%] & 12.8\% [9.9\%, 16.4\%] \\
Qwen2.5-3B & 409 & 23.2\% [19.4\%, 27.6\%] & 33.3\% [28.9\%, 38.0\%] & 30.8\% [26.5\%, 35.4\%] & 9.8\% [7.3\%, 13.0\%] \\
Qwen2.5-3B-I & 426 & 25.6\% [21.7\%, 29.9\%] & 37.8\% [33.3\%, 42.5\%] & 30.3\% [26.1\%, 34.8\%] & 10.8\% [8.2\%, 14.1\%] \\
Qwen2.5-7B & 403 & 25.8\% [21.8\%, 30.3\%] & 40.2\% [35.5\%, 45.1\%] & 29.3\% [25.0\%, 33.9\%] & 12.2\% [9.3\%, 15.7\%] \\
Mistral-7B & 369 & 38.5\% [33.7\%, 43.5\%] & 41.7\% [36.8\%, 46.8\%] & 30.6\% [26.1\%, 35.5\%] & 16.0\% [12.6\%, 20.1\%] \\
\bottomrule\end{tabular}\end{table}

\begin{table}[!htbp]\centering\color{revision}
\caption{Trained versus randomized-label probe on failures, and Jaccard overlaps between readable sets. Differences are percentage points with bootstrap 95\% intervals; all paired exact tests give $p<0.001$. L, T and P denote the logit, tuned and trained-probe readers.}\label{tab:probecontrol}
\footnotesize
\setlength{\tabcolsep}{4pt}
\begin{tabular}{lrrrrrrr}
\toprule
Model & $n_F$ & Trained only & Shuffled only & Difference [95\% CI] & L/T & L/P & T/P \\
\midrule
Llama-3.1-8B & 358 & 91 & 28 & 17.6\% [11.7\%, 23.2\%] & 0.471 & 0.288 & 0.469 \\
Llama-3.2-1B & 423 & 99 & 39 & 14.2\% [9.0\%, 19.4\%] & 0.397 & 0.251 & 0.431 \\
Llama-3.2-3B & 383 & 86 & 33 & 13.8\% [8.6\%, 18.8\%] & 0.587 & 0.300 & 0.437 \\
Llama-3.2-3B-I & 399 & 97 & 33 & 16.0\% [10.8\%, 21.3\%] & 0.530 & 0.300 & 0.374 \\
Qwen2.5-3B & 409 & 109 & 29 & 19.6\% [14.4\%, 24.7\%] & 0.383 & 0.277 & 0.456 \\
Qwen2.5-3B-I & 426 & 111 & 27 & 19.7\% [14.6\%, 24.4\%] & 0.357 & 0.293 & 0.443 \\
Qwen2.5-7B & 403 & 92 & 34 & 14.4\% [9.2\%, 19.4\%] & 0.378 & 0.354 & 0.451 \\
Mistral-7B & 369 & 98 & 22 & 20.6\% [15.4\%, 26.0\%] & 0.495 & 0.321 & 0.500 \\
\bottomrule\end{tabular}\end{table}

The gold token is hard-readable under a reader with scores $s_\ell$ when
\begin{equation}
 \max_{\ell\in\mathcal B}s_\ell(h_i^{(\ell)},a_i)>\max_{d\in D_i}\max_{\ell\in\mathcal B}s_\ell(h_i^{(\ell)},d),\qquad \mathcal B=\{\ell:0.4\leq\ell/L\leq0.9\},
\end{equation}
where $D_i$ is the decoy set of the item. \rev{The randomized-label control tests whether the candidate sets and the training procedure alone produce readable flags at the trained probe's rate; it is a control on supervision, not a permutation of labels across facts.}
\FloatBarrier

\section{Projection Interventions}
\label{app:interventions}
For unit output directions $u_a$ and $u_b$ the edits are $A(h)=h+(T_a-u_a^\top h)u_a$ and $B(h)=h+(T_b-u_b^\top h)u_b$, with $T_a$ and $T_b$ the corresponding percentiles or means of answer and alternative support among successes; \rev{the clamped variant used in Appendix~\ref{app:more} is $B{\downarrow}(h)=h+(\min(u_b^\top h,T_b)-u_b^\top h)u_b$}. The combined arm is $B(A(h))$, so the answer projection after both steps is $T_a+(T_b-u_b^\top A(h))\,u_a^\top u_b$, which differs from $T_a$ only through the cosine between the two directions; \rev{with the symmetric B this can lower the combined arm below A at high targets, as the p75 rows of Table~\ref{tab:allsweeps} show for two models, \rev{and clamping prevents upward alternative edits without guaranteeing that the combined arm outperforms A}}. Random controls have norm $|T_a-u_a^\top h|$.

\begin{table}[!htbp]\centering\color{revision}
\caption{Offline first-token recovery across calibration targets. Brackets are 95\% Wilson intervals.}\label{tab:allsweeps}
\footnotesize
\setlength{\tabcolsep}{4pt}
\begin{tabular}{lrrrrr}
\toprule
Model & Target & A & B & A$\to$B & Random \\
\midrule
Llama-3.1-8B & p25 & 63.7\% [58.6\%, 68.5\%] & 10.1\% [7.4\%, 13.6\%] & 74.9\% [70.1\%, 79.1\%] & 0.6\% [0.2\%, 2.0\%] \\
 & p50 & 84.9\% [80.8\%, 88.3\%] & 7.3\% [5.0\%, 10.4\%] & 90.5\% [87.0\%, 93.1\%] & 0.6\% [0.2\%, 2.0\%] \\
 & p75 & 94.4\% [91.5\%, 96.4\%] & 3.4\% [1.9\%, 5.8\%] & 97.8\% [95.7\%, 98.9\%] & 0.3\% [0.0\%, 1.6\%] \\
 & mean & 86.0\% [82.1\%, 89.2\%] & 7.3\% [5.0\%, 10.4\%] & 92.5\% [89.2\%, 94.8\%] & 0.6\% [0.2\%, 2.0\%] \\
Llama-3.2-1B & p25 & 31.9\% [27.7\%, 36.5\%] & 7.3\% [5.2\%, 10.2\%] & 44.0\% [39.3\%, 48.7\%] & 0.2\% [0.0\%, 1.3\%] \\
 & p50 & 83.0\% [79.1\%, 86.3\%] & 4.3\% [2.7\%, 6.6\%] & 85.6\% [81.9\%, 88.6\%] & 0.5\% [0.1\%, 1.7\%] \\
 & p75 & 96.5\% [94.2\%, 97.8\%] & 1.2\% [0.5\%, 2.7\%] & 94.1\% [91.4\%, 96.0\%] & 0.7\% [0.2\%, 2.1\%] \\
 & mean & 79.0\% [74.8\%, 82.6\%] & 5.4\% [3.7\%, 8.0\%] & 82.7\% [78.9\%, 86.0\%] & 0.2\% [0.0\%, 1.3\%] \\
Llama-3.2-3B & p25 & 64.0\% [59.0\%, 68.6\%] & 6.8\% [4.7\%, 9.8\%] & 76.0\% [71.5\%, 80.0\%] & 0.3\% [0.0\%, 1.5\%] \\
 & p50 & 83.0\% [78.9\%, 86.5\%] & 5.2\% [3.4\%, 7.9\%] & 90.9\% [87.6\%, 93.4\%] & 0.3\% [0.0\%, 1.5\%] \\
 & p75 & 96.1\% [93.6\%, 97.6\%] & 1.8\% [0.9\%, 3.7\%] & 96.6\% [94.3\%, 98.0\%] & 0.5\% [0.1\%, 1.9\%] \\
 & mean & 85.6\% [81.8\%, 88.8\%] & 5.2\% [3.4\%, 7.9\%] & 93.5\% [90.5\%, 95.5\%] & 0.3\% [0.0\%, 1.5\%] \\
Llama-3.2-3B-I & p25 & 81.7\% [77.6\%, 85.2\%] & 4.8\% [3.1\%, 7.3\%] & 92.0\% [88.9\%, 94.3\%] & 0.0\% [0.0\%, 1.0\%] \\
 & p50 & 92.2\% [89.2\%, 94.5\%] & 3.5\% [2.1\%, 5.8\%] & 96.7\% [94.5\%, 98.1\%] & 0.0\% [0.0\%, 1.0\%] \\
 & p75 & 97.2\% [95.1\%, 98.5\%] & 1.8\% [0.9\%, 3.6\%] & 99.7\% [98.6\%, 100.0\%] & 0.0\% [0.0\%, 1.0\%] \\
 & mean & 92.7\% [89.8\%, 94.9\%] & 3.3\% [1.9\%, 5.5\%] & 96.7\% [94.5\%, 98.1\%] & 0.0\% [0.0\%, 1.0\%] \\
Qwen2.5-3B & p25 & 80.9\% [76.8\%, 84.4\%] & 5.9\% [4.0\%, 8.6\%] & 87.8\% [84.2\%, 90.6\%] & 0.2\% [0.0\%, 1.4\%] \\
 & p50 & 90.7\% [87.5\%, 93.2\%] & 2.9\% [1.7\%, 5.1\%] & 94.6\% [92.0\%, 96.4\%] & 0.5\% [0.1\%, 1.8\%] \\
 & p75 & 97.1\% [94.9\%, 98.3\%] & 2.2\% [1.2\%, 4.1\%] & 99.5\% [98.2\%, 99.9\%] & 0.2\% [0.0\%, 1.4\%] \\
 & mean & 92.2\% [89.2\%, 94.4\%] & 3.2\% [1.9\%, 5.4\%] & 98.0\% [96.2\%, 99.0\%] & 0.7\% [0.2\%, 2.1\%] \\
Qwen2.5-3B-I & p25 & 76.5\% [72.3\%, 80.3\%] & 3.8\% [2.3\%, 6.0\%] & 86.2\% [82.5\%, 89.1\%] & 0.2\% [0.0\%, 1.3\%] \\
 & p50 & 90.6\% [87.5\%, 93.0\%] & 2.6\% [1.4\%, 4.6\%] & 95.3\% [92.9\%, 96.9\%] & 0.0\% [0.0\%, 0.9\%] \\
 & p75 & 98.4\% [96.6\%, 99.2\%] & 0.9\% [0.4\%, 2.4\%] & 99.5\% [98.3\%, 99.9\%] & 0.7\% [0.2\%, 2.0\%] \\
 & mean & 92.5\% [89.6\%, 94.6\%] & 2.6\% [1.4\%, 4.6\%] & 97.2\% [95.1\%, 98.4\%] & 0.0\% [0.0\%, 0.9\%] \\
Qwen2.5-7B & p25 & 79.2\% [74.9\%, 82.8\%] & 5.7\% [3.8\%, 8.4\%] & 84.1\% [80.2\%, 87.4\%] & 0.2\% [0.0\%, 1.4\%] \\
 & p50 & 89.3\% [85.9\%, 92.0\%] & 3.2\% [1.9\%, 5.4\%] & 90.6\% [87.3\%, 93.1\%] & 0.2\% [0.0\%, 1.4\%] \\
 & p75 & 95.3\% [92.8\%, 97.0\%] & 0.5\% [0.1\%, 1.8\%] & 93.5\% [90.7\%, 95.6\%] & 0.0\% [0.0\%, 0.9\%] \\
 & mean & 91.6\% [88.4\%, 93.9\%] & 3.2\% [1.9\%, 5.4\%] & 92.8\% [89.9\%, 94.9\%] & 0.2\% [0.0\%, 1.4\%] \\
Mistral-7B & p25 & 71.8\% [67.0\%, 76.2\%] & 10.0\% [7.4\%, 13.5\%] & 86.7\% [82.9\%, 89.8\%] & 0.0\% [0.0\%, 1.0\%] \\
 & p50 & 87.0\% [83.2\%, 90.0\%] & 7.6\% [5.3\%, 10.7\%] & 94.9\% [92.1\%, 96.7\%] & 0.0\% [0.0\%, 1.0\%] \\
 & p75 & 96.2\% [93.7\%, 97.7\%] & 3.8\% [2.3\%, 6.3\%] & 97.3\% [95.1\%, 98.5\%] & 0.8\% [0.3\%, 2.4\%] \\
 & mean & 87.3\% [83.5\%, 90.3\%] & 7.3\% [5.1\%, 10.4\%] & 94.3\% [91.5\%, 96.2\%] & 0.5\% [0.1\%, 2.0\%] \\
\bottomrule\end{tabular}\end{table}

\begin{table}[!htbp]\centering\color{revision}
\caption{Combined minus answer-only recovery at the median target, paired by question. Differences are percentage points with bootstrap 95\% intervals; Holm correction covers the four targets within each model.}\label{tab:incremental}
\small
\setlength{\tabcolsep}{4pt}
\begin{tabular}{lrrr}
\toprule
Model & $n$ & Paired difference [95\% CI] & $p_{\mathrm{Holm}}$ \\
\midrule
Llama-3.1-8B & 355 & 5.1\% [1.7\%, 8.7\%] & 0.013 \\
Llama-3.2-1B & 419 & 2.6\% [$-0.5$\%, 5.7\%] & 0.135 \\
Llama-3.2-3B & 380 & 7.6\% [4.5\%, 10.8\%] & $<0.001$ \\
Llama-3.2-3B-I & 396 & 4.5\% [2.5\%, 6.8\%] & $<0.001$ \\
Qwen2.5-3B & 406 & 3.9\% [1.5\%, 6.7\%] & 0.013 \\
Qwen2.5-3B-I & 423 & 4.7\% [2.6\%, 6.9\%] & $<0.001$ \\
Qwen2.5-7B & 401 & 1.0\% [$-1.5$\%, 3.7\%] & 0.884 \\
Mistral-7B & 366 & 7.9\% [4.6\%, 11.2\%] & $<0.001$ \\
\bottomrule\end{tabular}\end{table}
\FloatBarrier

\section{Reader-Stratified and Matched Recovery}
\label{app:matched}
\begin{table}[!htbp]\centering\color{revision}
\caption{Recovery under the median answer-setting edit on the readable subset of each reader, with subset sizes and 95\% Wilson intervals.}\label{tab:allstrata}
\scriptsize
\setlength{\tabcolsep}{4pt}
\begin{tabular}{lrrrrrrrr}
\toprule
Model & $n_L$ & Logit & $n_T$ & Tuned & $n_P$ & Probe & $n_C$ & Consensus \\
\midrule
Llama-3.1-8B & 129 & 75.2\% [67.1\%, 81.8\%] & 151 & 77.5\% [70.2\%, 83.4\%] & 108 & 75.0\% [66.1\%, 82.2\%] & 47 & 70.2\% [56.0\%, 81.3\%] \\
Llama-3.2-1B & 114 & 76.3\% [67.7\%, 83.2\%] & 164 & 75.6\% [68.5\%, 81.5\%] & 119 & 73.9\% [65.4\%, 81.0\%] & 44 & 68.2\% [53.4\%, 80.0\%] \\
Llama-3.2-3B & 161 & 80.1\% [73.3\%, 85.6\%] & 159 & 77.4\% [70.3\%, 83.2\%] & 104 & 74.0\% [64.9\%, 81.5\%] & 58 & 77.6\% [65.3\%, 86.4\%] \\
Llama-3.2-3B-I & 148 & 87.2\% [80.8\%, 91.6\%] & 157 & 89.8\% [84.1\%, 93.6\%] & 112 & 90.2\% [83.3\%, 94.4\%] & 51 & 86.3\% [74.3\%, 93.2\%] \\
Qwen2.5-3B & 95 & 83.2\% [74.4\%, 89.4\%] & 136 & 85.3\% [78.4\%, 90.3\%] & 126 & 88.1\% [81.3\%, 92.7\%] & 40 & 82.5\% [68.0\%, 91.3\%] \\
Qwen2.5-3B-I & 108 & 85.2\% [77.3\%, 90.7\%] & 162 & 86.4\% [80.3\%, 90.9\%] & 130 & 90.0\% [83.6\%, 94.1\%] & 46 & 91.3\% [79.7\%, 96.6\%] \\
Qwen2.5-7B & 104 & 76.9\% [68.0\%, 84.0\%] & 162 & 80.2\% [73.4\%, 85.6\%] & 119 & 81.5\% [73.6\%, 87.5\%] & 49 & 67.3\% [53.4\%, 78.8\%] \\
Mistral-7B & 142 & 81.0\% [73.7\%, 86.6\%] & 154 & 78.6\% [71.4\%, 84.3\%] & 114 & 78.9\% [70.6\%, 85.4\%] & 60 & 75.0\% [62.8\%, 84.2\%] \\
\bottomrule\end{tabular}\end{table}

\begin{table}[!htbp]\centering\color{revision}
\caption{Readable (R) versus unreadable (U) failures matched on the initial margin, for the logit and probe readers. Differences are R minus U with bootstrap 95\% intervals; gap is the mean absolute margin mismatch of the pairs.}\label{tab:allmatched}
\footnotesize
\setlength{\tabcolsep}{4pt}
\begin{tabular}{lrrrrrr}
\toprule
Model & Reader & $n$ & R / U & Difference [95\% CI] & Edit-norm difference [95\% CI] & Gap \\
\midrule
Llama-3.1-8B & logit & 129 & 75.2\% / 91.5\% & $-16.3$\% [$-25.6$\%, $-7.8$\%] & $-2.05$ [$-2.62$, $-1.48$] & 0.553 \\
 & probe & 108 & 75.0\% / 88.0\% & $-13.0$\% [$-22.2$\%, $-3.7$\%] & $-1.34$ [$-1.91$, $-0.77$] & 0.123 \\
Llama-3.2-1B & logit & 114 & 76.3\% / 83.3\% & $-7.0$\% [$-18.4$\%, 3.5\%] & $-0.79$ [$-1.31$, $-0.24$] & 0.360 \\
 & probe & 119 & 73.9\% / 83.2\% & $-9.2$\% [$-20.2$\%, 1.7\%] & $-0.77$ [$-1.26$, $-0.30$] & 0.336 \\
Llama-3.2-3B & logit & 161 & 80.1\% / 87.0\% & $-6.8$\% [$-14.9$\%, 1.2\%] & $-1.55$ [$-1.90$, $-1.22$] & 1.337 \\
 & probe & 104 & 74.0\% / 89.4\% & $-15.4$\% [$-25.0$\%, $-5.8$\%] & $-0.71$ [$-1.00$, $-0.42$] & 0.048 \\
Llama-3.2-3B-I & logit & 148 & 87.2\% / 96.6\% & $-9.5$\% [$-15.5$\%, $-3.4$\%] & $-2.09$ [$-2.54$, $-1.65$] & 0.971 \\
 & probe & 112 & 90.2\% / 96.4\% & $-6.3$\% [$-13.4$\%, 0.9\%] & $-0.83$ [$-1.33$, $-0.33$] & 0.091 \\
Qwen2.5-3B & logit & 95 & 83.2\% / 91.6\% & $-8.4$\% [$-16.8$\%, 0.0\%] & $-0.95$ [$-1.40$, $-0.47$] & 0.128 \\
 & probe & 126 & 88.1\% / 92.9\% & $-4.8$\% [$-11.9$\%, 2.4\%] & $-1.18$ [$-1.54$, $-0.81$] & 0.330 \\
Qwen2.5-3B-I & logit & 108 & 85.2\% / 93.5\% & $-8.3$\% [$-16.7$\%, $-0.9$\%] & $-1.71$ [$-2.59$, $-0.94$] & 0.052 \\
 & probe & 130 & 90.0\% / 91.5\% & $-1.5$\% [$-8.5$\%, 5.4\%] & $-1.31$ [$-2.04$, $-0.60$] & 0.168 \\
Qwen2.5-7B & logit & 104 & 76.9\% / 92.3\% & $-15.4$\% [$-24.0$\%, $-6.7$\%] & $-1.57$ [$-2.28$, $-0.91$] & 0.085 \\
 & probe & 119 & 81.5\% / 91.6\% & $-10.1$\% [$-18.5$\%, $-1.7$\%] & $-1.93$ [$-2.51$, $-1.27$] & 0.266 \\
Mistral-7B & logit & 142 & 81.0\% / 91.5\% & $-10.6$\% [$-18.3$\%, $-2.8$\%] & $-5.71$ [$-7.35$, $-4.16$] & 0.857 \\
 & probe & 114 & 78.9\% / 91.2\% & $-12.3$\% [$-21.9$\%, $-3.5$\%] & $-3.92$ [$-5.41$, $-2.42$] & 0.166 \\
\bottomrule\end{tabular}\end{table}

\begin{revision}
Each readable failure is paired with the unused unreadable failure closest in absolute initial margin, in record order and without a caliper, and the pairs are not additionally balanced on relation, token identity, frequency or answer support; the mean margin gap of the pairs is reported so that the quality of the match can be judged. The comparison is observational and does not establish a causal effect of readability on recoverability; \rev{smaller edit norms indicate less distance to the calibration target, not necessarily larger answer or alternative logits.}
\end{revision}
\FloatBarrier

\section{Component and Residual Transfer Diagnostics}
\label{app:components}
\begin{revision}
All transfers replace last-position activations. The late band begins at $\lfloor0.75L\rfloor$; replacing every attention and MLP output in it fixes the band's additive contributions, after which only the final normalization acts, whereas replacing only the attention or only the MLP outputs, or the residual entering the band, leaves the other components free to respond. The early and middle bands have the same number of components as the late band, and their replacement leaves all later layers unpatched. The different-fact donor is the successful variant of another pair, not matched for relation or answer, and the random-component control samples the same number of components across all depths, so that it varies both the location and the amount of constrained computation.
\end{revision}

\begin{table}[!htbp]\centering\color{revision}
\caption{First-token recovery under component replacement. SF and DF replace every late attention and MLP output with same-fact and different-fact donor outputs; Random replaces the same number of components sampled across all depths; Attn and MLP replace only the late attention or only the late MLP outputs, with same-fact and different-fact (DF) donors; Middle and Early replace same-size bands.}\label{tab:allpatch}
\footnotesize
\setlength{\tabcolsep}{4pt}
\begin{tabular}{lrrrrrrrrrr}
\toprule
Model & $n$ & SF & DF & Random & Attn & Attn DF & MLP & MLP DF & Middle & Early \\
\midrule
Llama-3.1-8B & 45 & 75.6\% & 6.7\% & 77.8\% & 24.4\% & 22.2\% & 64.4\% & 8.9\% & 62.2\% & 37.8\% \\
Llama-3.2-1B & 28 & 75.0\% & 3.6\% & 42.9\% & 46.4\% & 32.1\% & 50.0\% & 3.6\% & 53.6\% & 21.4\% \\
Llama-3.2-3B & 31 & 74.2\% & 6.5\% & 77.4\% & 35.5\% & 35.5\% & 61.3\% & 9.7\% & 74.2\% & 25.8\% \\
Llama-3.2-3B-I & 22 & 68.2\% & 0.0\% & 59.1\% & 31.8\% & 22.7\% & 59.1\% & 4.5\% & 72.7\% & 27.3\% \\
Qwen2.5-3B & 18 & 83.3\% & 5.6\% & 50.0\% & 55.6\% & 11.1\% & 55.6\% & 0.0\% & 38.9\% & 22.2\% \\
Qwen2.5-3B-I & 25 & 100.0\% & 0.0\% & 60.0\% & 64.0\% & 0.0\% & 84.0\% & 0.0\% & 48.0\% & 32.0\% \\
Qwen2.5-7B & 20 & 90.0\% & 0.0\% & 80.0\% & 45.0\% & 0.0\% & 90.0\% & 0.0\% & 45.0\% & 45.0\% \\
Mistral-7B & 34 & 64.7\% & 0.0\% & 55.9\% & 35.3\% & 32.4\% & 61.8\% & 5.9\% & 67.6\% & 20.6\% \\
\bottomrule\end{tabular}\end{table}

\begin{table}[!htbp]\centering\color{revision}
\caption{Paired \rev{component-replacement comparisons paired by question string.} Differences are percentage points with bootstrap 95\% intervals; Holm correction covers the nine recovery comparisons per model.}\label{tab:patchcomparisons}
\footnotesize
\setlength{\tabcolsep}{4pt}
\begin{tabular}{lrrrr}
\toprule
Model & Comparison & $n$ & Difference [95\% CI] & $p_{\mathrm{Holm}}$ \\
\midrule
Llama-3.1-8B & SF vs random & 44 & $-2.3$\% [$-20.5$\%, 13.6\%] & 1.000 \\
 & Attn vs DF & 44 & 2.3\% [$-13.6$\%, 18.2\%] & 1.000 \\
 & MLP vs DF & 44 & 54.5\% [38.6\%, 70.5\%] & $<0.001$ \\
Llama-3.2-1B & SF vs random & 27 & 33.3\% [14.8\%, 55.6\%] & 0.059 \\
 & Attn vs DF & 27 & 14.8\% [$-7.4$\%, 37.0\%] & 1.000 \\
 & MLP vs DF & 27 & 44.4\% [22.2\%, 66.7\%] & 0.011 \\
Llama-3.2-3B & SF vs random & 31 & $-3.2$\% [$-19.4$\%, 12.9\%] & 1.000 \\
 & Attn vs DF & 31 & 0.0\% [$-19.4$\%, 19.4\%] & 1.000 \\
 & MLP vs DF & 31 & 51.6\% [29.0\%, 71.0\%] & 0.003 \\
Llama-3.2-3B-I & SF vs random & 22 & 9.1\% [$-22.7$\%, 40.9\%] & 1.000 \\
 & Attn vs DF & 22 & 9.1\% [$-13.6$\%, 31.8\%] & 1.000 \\
 & MLP vs DF & 22 & 54.5\% [31.8\%, 77.3\%] & 0.013 \\
Qwen2.5-3B & SF vs random & 18 & 33.3\% [5.6\%, 61.1\%] & 0.211 \\
 & Attn vs DF & 18 & 44.4\% [11.1\%, 72.2\%] & 0.086 \\
 & MLP vs DF & 18 & 55.6\% [33.3\%, 77.8\%] & 0.014 \\
Qwen2.5-3B-I & SF vs random & 25 & 40.0\% [20.0\%, 60.0\%] & 0.006 \\
 & Attn vs DF & 25 & 64.0\% [44.0\%, 80.0\%] & $<0.001$ \\
 & MLP vs DF & 25 & 84.0\% [68.0\%, 96.0\%] & $<0.001$ \\
Qwen2.5-7B & SF vs random & 20 & 10.0\% [$-10.0$\%, 30.0\%] & 1.000 \\
 & Attn vs DF & 20 & 45.0\% [25.0\%, 70.0\%] & 0.023 \\
 & MLP vs DF & 20 & 90.0\% [75.0\%, 100.0\%] & $<0.001$ \\
Mistral-7B & SF vs random & 33 & 9.1\% [$-12.1$\%, 30.3\%] & 1.000 \\
 & Attn vs DF & 33 & 3.0\% [$-21.2$\%, 27.3\%] & 1.000 \\
 & MLP vs DF & 33 & 54.5\% [36.4\%, 69.7\%] & $<0.001$ \\
\bottomrule\end{tabular}\end{table}

\begin{table}[!htbp]\centering\color{revision}
\caption{Residual patching at the layer where the logit lens reads the gold token most strongly, from a same-fact donor against a different-fact donor and against the same donor at a random other band layer. \rev{Differences are percentage points, paired by question string.}}\label{tab:peak}
\footnotesize
\setlength{\tabcolsep}{4pt}
\begin{tabular}{lrrrr}
\toprule
Model & Control & $n$ & Recovery difference [95\% CI] & $p_{\mathrm{Holm}}$ \\
\midrule
Llama-3.1-8B & Different fact & 44 & 84.1\% [72.7\%, 93.2\%] & $<0.001$ \\
 & Random layer & 44 & 0.0\% [$-6.8$\%, 6.8\%] & 1.000 \\
Llama-3.2-1B & Different fact & 27 & 85.2\% [70.4\%, 96.3\%] & $<0.001$ \\
 & Random layer & 27 & 22.2\% [3.7\%, 40.7\%] & 0.070 \\
Llama-3.2-3B & Different fact & 31 & 77.4\% [58.1\%, 93.5\%] & $<0.001$ \\
 & Random layer & 31 & 3.2\% [$-12.9$\%, 19.4\%] & 1.000 \\
Llama-3.2-3B-I & Different fact & 22 & 81.8\% [63.6\%, 95.5\%] & $<0.001$ \\
 & Random layer & 22 & 9.1\% [$-4.5$\%, 27.3\%] & 0.625 \\
Qwen2.5-3B & Different fact & 18 & 27.8\% [0.0\%, 50.0\%] & 0.250 \\
 & Random layer & 18 & 11.1\% [0.0\%, 27.8\%] & 0.500 \\
Qwen2.5-3B-I & Different fact & 25 & 52.0\% [28.0\%, 72.0\%] & 0.002 \\
 & Random layer & 25 & 8.0\% [$-12.0$\%, 28.0\%] & 0.688 \\
Qwen2.5-7B & Different fact & 20 & 75.0\% [55.0\%, 95.0\%] & $<0.001$ \\
 & Random layer & 20 & 10.0\% [$-10.0$\%, 30.0\%] & 0.625 \\
Mistral-7B & Different fact & 33 & 72.7\% [57.6\%, 87.9\%] & $<0.001$ \\
 & Random layer & 33 & 18.2\% [3.0\%, 33.3\%] & 0.070 \\
\bottomrule\end{tabular}\end{table}

\begin{revision}
The peak layer is the measured one when the failing variant is the evaluation prompt and the centre of the read band otherwise. In the direction-ablation test the gold read direction is proportional to $g\odot W_U[a]$, with $g$ the weight of the final RMS normalization, and the random control subtracts the same signed projection magnitude along a random unit vector; \rev{the test selects the first 60 failures marked logit-readable through question-keyed flags. Subset membership and layer assignments remain unverified at the record level for these eight runs.}
\end{revision}

\begin{table}[!htbp]\centering\color{revision}
\caption{Ablation of the gold read direction at the peak-read layer on 60 logit-readable failures per model. Values are changes in the final answer-versus-alternative margin, with paired bootstrap 95\% intervals.}\label{tab:ablation}
\small
\setlength{\tabcolsep}{4pt}
\begin{tabular}{lrrrr}
\toprule
Model & $n$ & Read-direction shift & Random shift & Paired difference \\
\midrule
Llama-3.1-8B & 60 & $-2.449$ [$-2.847$, $-2.087$] & 0.002 [$-0.031$, 0.036] & $-2.452$ [$-2.850$, $-2.091$] \\
Llama-3.2-1B & 60 & $-3.157$ [$-3.532$, $-2.795$] & 0.007 [$-0.047$, 0.058] & $-3.164$ [$-3.549$, $-2.801$] \\
Llama-3.2-3B & 60 & $-3.076$ [$-3.611$, $-2.554$] & $-0.033$ [$-0.097$, 0.018] & $-3.043$ [$-3.560$, $-2.536$] \\
Llama-3.2-3B-I & 60 & $-3.362$ [$-4.002$, $-2.774$] & 0.024 [$-0.026$, 0.073] & $-3.387$ [$-4.036$, $-2.784$] \\
Qwen2.5-3B & 60 & $-1.598$ [$-2.059$, $-1.144$] & 0.041 [$-0.013$, 0.099] & $-1.639$ [$-2.114$, $-1.188$] \\
Qwen2.5-3B-I & 60 & $-3.528$ [$-4.552$, $-2.484$] & $-0.051$ [$-0.146$, 0.041] & $-3.477$ [$-4.522$, $-2.415$] \\
Qwen2.5-7B & 60 & $-1.420$ [$-1.814$, $-1.069$] & 0.017 [$-0.004$, 0.039] & $-1.437$ [$-1.823$, $-1.092$] \\
Mistral-7B & 60 & $-3.364$ [$-3.906$, $-2.845$] & $-0.003$ [$-0.040$, 0.034] & $-3.360$ [$-3.900$, $-2.847$] \\
\bottomrule\end{tabular}\end{table}
\FloatBarrier

\section{Baseline Reference Sensitivity}
\label{app:baseline}
\begin{revision}
The all-record reference averages the normalized final states of the 500 evaluation prompts, the failure-conditioned reference uses their failure subset, and the neutral reference uses the first 256 WikiText-103 test lines longer than 40 characters, fed as raw text. Token log frequencies use the first 8,000 qualifying training lines, truncated to 256 tokens each, with add-one smoothing.
\end{revision}

\begin{table}[!htbp]\centering\color{revision}
\caption{Baseline sign and frequency association on failures under three reference means. Brackets are 95\% intervals.}\label{tab:baselinerefs}
\footnotesize
\setlength{\tabcolsep}{4pt}
\begin{tabular}{lrrrr}
\toprule
Model & Reference & Negative [95\% CI] & Mean margin [95\% CI] & $r_{\mathrm{freq}}$ \\
\midrule
Llama-3.1-8B & All records & 67.0\% [62.0\%, 71.7\%] & $-1.242$ [$-1.526$, $-0.937$] & 0.567 \\
 & Failures & 68.2\% [63.2\%, 72.8\%] & $-1.272$ [$-1.557$, $-0.972$] & 0.554 \\
 & Neutral text & 65.4\% [60.3\%, 70.1\%] & $-0.951$ [$-1.262$, $-0.620$] & 0.508 \\
Llama-3.2-1B & All records & 77.8\% [73.6\%, 81.5\%] & $-1.679$ [$-1.923$, $-1.421$] & 0.568 \\
 & Failures & 79.0\% [74.8\%, 82.6\%] & $-1.804$ [$-2.055$, $-1.538$] & 0.558 \\
 & Neutral text & 73.0\% [68.6\%, 77.1\%] & $-1.311$ [$-1.576$, $-1.038$] & 0.530 \\
Llama-3.2-3B & All records & 71.3\% [66.6\%, 75.6\%] & $-1.335$ [$-1.577$, $-1.098$] & 0.570 \\
 & Failures & 71.8\% [67.1\%, 76.1\%] & $-1.406$ [$-1.651$, $-1.165$] & 0.556 \\
 & Neutral text & 64.5\% [59.6\%, 69.1\%] & $-0.859$ [$-1.134$, $-0.573$] & 0.510 \\
Llama-3.2-3B-I & All records & 69.9\% [65.3\%, 74.2\%] & $-1.053$ [$-1.293$, $-0.810$] & 0.554 \\
 & Failures & 71.4\% [66.8\%, 75.6\%] & $-1.130$ [$-1.373$, $-0.882$] & 0.540 \\
 & Neutral text & 64.9\% [60.1\%, 69.4\%] & $-0.792$ [$-1.069$, $-0.520$] & 0.437 \\
Qwen2.5-3B & All records & 76.8\% [72.4\%, 80.6\%] & $-1.828$ [$-2.099$, $-1.552$] & 0.561 \\
 & Failures & 77.8\% [73.5\%, 81.5\%] & $-1.919$ [$-2.194$, $-1.641$] & 0.548 \\
 & Neutral text & 67.2\% [62.5\%, 71.6\%] & $-1.070$ [$-1.402$, $-0.774$] & 0.490 \\
Qwen2.5-3B-I & All records & 67.6\% [63.0\%, 71.9\%] & $-1.519$ [$-1.864$, $-1.170$] & 0.503 \\
 & Failures & 67.1\% [62.5\%, 71.4\%] & $-1.563$ [$-1.907$, $-1.209$] & 0.493 \\
 & Neutral text & 63.8\% [59.2\%, 68.3\%] & $-0.910$ [$-1.209$, $-0.599$] & 0.488 \\
Qwen2.5-7B & All records & 82.1\% [78.1\%, 85.6\%] & $-2.026$ [$-2.274$, $-1.783$] & 0.495 \\
 & Failures & 83.6\% [79.7\%, 86.9\%] & $-2.120$ [$-2.373$, $-1.868$] & 0.479 \\
 & Neutral text & 75.2\% [70.7\%, 79.2\%] & $-1.630$ [$-1.913$, $-1.357$] & 0.454 \\
Mistral-7B & All records & 66.9\% [62.0\%, 71.5\%] & $-1.141$ [$-1.408$, $-0.873$] & 0.500 \\
 & Failures & 67.8\% [62.8\%, 72.3\%] & $-1.211$ [$-1.483$, $-0.942$] & 0.488 \\
 & Neutral text & 67.2\% [62.3\%, 71.8\%] & $-1.119$ [$-1.442$, $-0.804$] & 0.438 \\
\bottomrule\end{tabular}\end{table}
\FloatBarrier

\section{Continuation Diagnostics}
\label{app:continuation}
The generated interventions edit the final normalized state at the first decoding step, cast it back to the model's activation dtype, and apply the first-step content mask; subsequent steps are unedited. Each generated item falls into one of four categories: N, the canonical first token is not recovered; A, it is recovered and the full answer is correct; D, it is recovered, the full answer is wrong, and the token is shared with another entity in the relation pool, so that the continuation may have completed a different entity with the same prefix; and B, the remaining wrong continuations after recovery of a unique prefix. Full-answer correctness can exceed category A because an item in category N can still generate an accepted alias that begins with a different token, and the difference between first-token recovery and full-answer correctness is therefore not a conditional probability of continuation error. \rev{The mask-only control, which generates with the first-step content mask and no edit, gives the full-answer rate attributable to the mask alone; it is reported for the Qwen3 checkpoints in Appendix~\ref{app:more}.}

\begin{revision}
\section{Additional Checkpoints: Qwen3-Base, 0.6B to 14B}
\label{app:more}
The entire protocol was repeated on \texttt{Qwen3-0.6B-Base}, \texttt{Qwen3-1.7B-Base}, \texttt{Qwen3-4B-Base}, \texttt{Qwen3-8B-Base} and \texttt{Qwen3-14B-Base}, with the same data split, prompts, demonstrations, seed and sample sizes as the main runs, and with the refinements described in Section~\ref{sec:methodology}: every join keyed on the record identifier, the clamped alternative edit B$\downarrow$, a mask-only generation control, and per-record reconstruction diagnostics. The pair search was allowed to visit more records, which yields 60 pairs per model. Four of the 500 evaluation records are repeated question strings in this split.

Across the family the dissociation is unchanged, with 26.8--33.4\% of failures readable under the logit lens, 33.6--40.7\% under the tuned reader and 25.1--27.7\% under the trained probe against 11.5--13.5\% for the randomized-label control, and the baseline is negative for 75--85\% of failures with the same correlation with token frequency as in the main runs (Table~\ref{tab:qwen3a}). What changes with scale is the share of failures that answer support alone can repair: it rises from 47.0\% at 0.6B to 85.4\% at 14B, while the share of failures with positive alternative excess falls from 50.7\% to 14.3\%, the excess criterion keeps zero false positives throughout, and the combined edit adds 4.4--20.3 points with every increment significant (Table~\ref{tab:qwen3b}). The mask-only control shows that the first-step content mask alone yields 1.5--5.0\% full-answer correctness on the same failures, against 20.7--37.1\% with the answer edit. \rev{The matched comparison directly records the projections and excess shares discussed in Section~\ref{sec:results}:} readable failures have higher answer support at matched margin in every model, by 0.74--1.69 units with all intervals above zero, \rev{the intervals for their alternative-projection differences include zero without establishing equivalence, their alternative-limited share is approximately 1.2--2.5 times} that of the matched unreadable failures, \rev{and their recovery point estimates are lower by 7.9--30.8 points, with intervals excluding zero in three of five models (Table~\ref{tab:qwen3c}).} The within-fact transfers give the same ordering as in the main runs, residual transfer and late MLPs above their different-fact controls in every model, late attention significant in two of five, \rev{the middle band also recovering many failures}, and the read-direction ablation lowering the final margin by 1.6--3.8 logits against a random direction (Table~\ref{tab:qwen3d}). The doubly normalized extraction of the submitted version is especially damaging in this family: its top-1 agreement with the native logits is 0.00--0.34, against 0.98--1.00 for the corrected extraction, with a maximum absolute logit error of 0.06--0.12 and every remaining top-1 disagreement an exact tie in the native logits.
\end{revision}

\begin{table}[!htbp]\centering\color{revision}
\caption{Qwen3-Base: protocol, readers and baseline on failures. Gate is the first-token protocol check among fully correct generations; reconstruction is top-1 agreement with the native logits for the corrected (new) and doubly normalized (old) extraction; reader columns are hard-decoy readable fractions among failures; probe$-$shuffled is the paired difference in percentage points with a bootstrap 95\% interval (all exact $p<0.001$); negative baseline is reported under the all-record and neutral-text references; frequency recovery is first-token recovery after removing the frequency direction, with the random control.}\label{tab:qwen3a}
\scriptsize
\setlength{\tabcolsep}{3pt}
\resizebox{\textwidth}{!}{%
\begin{tabular}{lrrrrrrrrrrrrrr}
\toprule
Model & Gate & Full acc & $n_F$ & Recon new/old & Logit & Tuned & Probe & Shuffled & Probe$-$shuffled & Neg. (all) & Neg. (neutral) & $r_{\mathrm{freq}}$ & Freq. rec. & Random \\
\midrule
Qwen3-0.6B & 94.7\% & 11.4\% & 434 & 1.00 / 0.34 & 27.9\% & 33.6\% & 26.5\% & 12.0\% & $+14.5$ [9.4, 19.4] & 84.6\% & 77.2\% & 0.558 & 0.9\% & 1.6\% \\
Qwen3-1.7B & 100.0\% & 14.8\% & 411 & 0.98 / 0.12 & 29.0\% & 35.3\% & 25.1\% & 12.7\% & $+12.4$ [7.5, 17.5] & 85.2\% & 74.0\% & 0.566 & 1.0\% & 1.5\% \\
Qwen3-4B & 98.9\% & 17.4\% & 401 & 1.00 / 0.02 & 29.2\% & 39.2\% & 27.7\% & 11.5\% & $+16.2$ [10.7, 21.7] & 80.8\% & 72.3\% & 0.564 & 0.7\% & 2.7\% \\
Qwen3-8B & 98.1\% & 21.4\% & 388 & 1.00 / 0.00 & 26.8\% & 40.7\% & 26.3\% & 12.6\% & $+13.7$ [8.5, 18.6] & 76.8\% & 69.8\% & 0.568 & 0.8\% & 1.5\% \\
Qwen3-14B & 96.0\% & 25.2\% & 377 & 1.00 / 0.26 & 33.4\% & 38.7\% & 27.6\% & 13.5\% & $+14.1$ [8.7, 19.4] & 75.3\% & 70.8\% & 0.561 & 1.1\% & 2.7\% \\
\bottomrule\end{tabular}}\end{table}

\begin{table}[!htbp]\centering\color{revision}
\caption{Qwen3-Base: median-calibrated edits on all failures. A sets the answer support, B$\downarrow$ lowers the alternative support to at most the median of successes, A$\to$B$\downarrow$ applies both, and Random matches the norm of A; combined$-$A is the paired difference in percentage points (bootstrap 95\% interval, Holm-adjusted exact $p<0.001$ in every model); excess $>0$ is the share of failures the criterion of equation~\ref{eq:excess} predicts answer setting cannot repair, with its false positives and misses; the last four columns are generated outcomes, with the mask-only control (first-step content mask, no edit).}\label{tab:qwen3b}
\scriptsize
\setlength{\tabcolsep}{3pt}
\resizebox{\textwidth}{!}{%
\begin{tabular}{lrrrrrrrrrrrr}
\toprule
Model & A & B$\downarrow$ & A$\to$B$\downarrow$ & Random & Combined$-$A & Excess $>0$ & FP / FN & A gen. first & A full & A$\to$B$\downarrow$ full & Mask only full & Offline $=$ gen. \\
\midrule
Qwen3-0.6B & 47.0\% & 1.2\% & 67.3\% & 0.0\% & $+20.3$ [16.6, 24.2] & 50.7\% & 0 / 10 & 46.8\% & 20.7\% & 25.6\% & 3.7\% & 0.970 \\
Qwen3-1.7B & 56.9\% & 1.9\% & 67.6\% & 0.5\% & $+10.7$ [7.8, 13.9] & 42.8\% & 0 / 1 & 56.2\% & 22.1\% & 26.3\% & 1.5\% & 0.981 \\
Qwen3-4B & 75.1\% & 2.7\% & 83.8\% & 0.0\% & $+8.7$ [6.0, 11.7] & 24.4\% & 0 / 2 & 75.3\% & 34.4\% & 36.2\% & 2.2\% & 0.993 \\
Qwen3-8B & 73.2\% & 5.9\% & 77.6\% & 0.8\% & $+4.4$ [2.3, 6.4] & 26.3\% & 0 / 2 & 72.9\% & 29.6\% & 33.2\% & 3.6\% & 0.992 \\
Qwen3-14B & 85.4\% & 5.0\% & 91.5\% & 0.5\% & $+6.1$ [3.7, 8.8] & 14.3\% & 0 / 1 & 85.4\% & 37.1\% & 40.3\% & 5.0\% & 0.989 \\
\bottomrule\end{tabular}}\end{table}

\begin{table}[!htbp]\centering\color{revision}
\caption{Qwen3-Base: readable (R) versus unreadable (U) failures under the logit reader, matched on the initial margin, at the median answer-setting edit. Differences are R minus U with bootstrap 95\% intervals; $\pi_a$ and $\pi_b$ are the projections on the answer and alternative directions before any edit; alternative-limited is the share of each subset with positive excess.}\label{tab:qwen3c}
\footnotesize
\setlength{\tabcolsep}{4pt}
\resizebox{\textwidth}{!}{%
\begin{tabular}{lrrrrrrr}
\toprule
Model & $n$ & R / U recovery & Difference & Edit-norm diff. & $\pi_a$ diff. & $\pi_b$ diff. & Alt-limited R / U \\
\midrule
Qwen3-0.6B & 121 & 31.4\% / 43.0\% & $-11.6$ [$-23.1$, 0.0] & $-0.93$ [$-1.33$, $-0.48$] & $+1.43$ [0.87, 1.96] & $-0.07$ [$-0.74$, 0.54] & 0.67 / 0.55 \\
Qwen3-1.7B & 119 & 31.9\% / 52.1\% & $-20.2$ [$-31.9$, $-7.6$] & $-0.60$ [$-0.89$, $-0.30$] & $+1.02$ [0.65, 1.40] & $+0.19$ [$-0.19$, 0.58] & 0.67 / 0.48 \\
Qwen3-4B & 117 & 49.6\% / 80.3\% & $-30.8$ [$-41.0$, $-20.5$] & $-1.48$ [$-1.90$, $-1.04$] & $+1.69$ [1.22, 2.15] & $+0.00$ [$-0.59$, 0.56] & 0.50 / 0.20 \\
Qwen3-8B & 104 & 49.0\% / 78.8\% & $-29.8$ [$-41.3$, $-18.3$] & $-0.63$ [$-1.06$, $-0.20$] & $+1.03$ [0.53, 1.52] & $-0.09$ [$-0.68$, 0.48] & 0.50 / 0.21 \\
Qwen3-14B & 126 & 77.8\% / 85.7\% & $-7.9$ [$-18.3$, 2.4] & $-0.60$ [$-1.03$, $-0.17$] & $+0.74$ [0.23, 1.24] & $-0.16$ [$-0.66$, 0.31] & 0.22 / 0.13 \\
\bottomrule\end{tabular}}\end{table}

\begin{table}[!htbp]\centering\color{revision}
\caption{Qwen3-Base: within-fact transfers on 60 pairs per model, first-token recovery for the same-fact donor / the different-fact donor, with Holm-adjusted exact $p$ for the paired difference; middle and early replace same-size bands; late-all replaces every late block (same fact / different fact / random components across depth); peak-layer is the paired recovery difference of same-fact over different-fact residual patching at the peak-read layer; ablation is the change in the final margin from removing the gold read direction minus that of a random direction.}\label{tab:qwen3d}
\scriptsize
\setlength{\tabcolsep}{3pt}
\resizebox{\textwidth}{!}{%
\begin{tabular}{lrrrrrrrrrrr}
\toprule
Model & Band input & $p$ & Late MLP & $p$ & Late attn & $p$ & Middle & Early & Late-all SF / DF / random & Peak-layer & Ablation \\
\midrule
Qwen3-0.6B & 90.0 / 23.3 & $<0.001$ & 75.0 / 31.7 & $<0.001$ & 36.7 / 21.7 & 0.216 & 70.0 & 41.7 & 83.3 / 23.3 / 66.7 & $+70.0$ [56.7, 81.7] & $-2.56$ [$-3.04$, $-2.10$] \\
Qwen3-1.7B & 83.3 / 6.7 & $<0.001$ & 58.3 / 11.7 & $<0.001$ & 33.3 / 21.7 & 0.568 & 76.7 & 50.0 & 71.7 / 5.0 / 65.0 & $+68.3$ [53.3, 81.7] & $-1.98$ [$-2.51$, $-1.48$] \\
Qwen3-4B & 80.0 / 3.3 & $<0.001$ & 65.0 / 6.7 & $<0.001$ & 45.0 / 15.0 & 0.004 & 53.3 & 36.7 & 86.7 / 1.7 / 63.3 & $+68.3$ [55.0, 80.0] & $-1.62$ [$-2.13$, $-1.16$] \\
Qwen3-8B & 73.3 / 1.7 & $<0.001$ & 56.7 / 1.7 & $<0.001$ & 31.7 / 23.3 & 1.000 & 51.7 & 28.3 & 58.3 / 3.3 / 51.7 & $+56.7$ [43.3, 70.0] & $-2.20$ [$-2.87$, $-1.55$] \\
Qwen3-14B & 66.7 / 0.0 & $<0.001$ & 58.3 / 0.0 & $<0.001$ & 56.7 / 15.0 & $<0.001$ & 51.7 & 21.7 & 83.3 / 0.0 / 51.7 & $+56.7$ [41.7, 70.0] & $-3.80$ [$-4.49$, $-3.13$] \\
\bottomrule\end{tabular}}\end{table}
\FloatBarrier

\end{document}

%% file: math_commands.tex
\usepackage{amsmath,amsfonts,bm}

\def\eqref#1{equation~\ref{#1}}

\def\1{\bm{1}}

\DeclareMathAlphabet{\mathsfit}{\encodingdefault}{\sfdefault}{m}{sl}
\SetMathAlphabet{\mathsfit}{bold}{\encodingdefault}{\sfdefault}{bx}{n}